%% file: main.tex
\documentclass{article}

\usepackage[nonatbib,final]{neurips_2021}

\usepackage[numbers]{natbib}
\usepackage[utf8]{inputenc} % allow utf-8 input
\usepackage[T1]{fontenc}    % use 8-bit T1 fonts
\usepackage{hyperref}       % hyperlinks
\usepackage{url}            % simple URL typesetting
\usepackage{booktabs}       % professional-quality tables
\usepackage{amsfonts}       % blackboard math symbols
\usepackage{nicefrac}       % compact symbols for 1/2, etc.
\usepackage{microtype}      % microtypography

\input{math_commands}
\usepackage{soul}
\usepackage[makeroom]{cancel}

\usepackage{amsmath,bm,amsthm}
\newtheorem{theorem}{Theorem}[section]
\newtheorem{lemma}{Lemma}[section]

\theoremstyle{remark}
\newtheorem{definition}{Definition}[section]
\newtheorem{condition}{Condition}[section]

\theoremstyle{remark}
\newtheorem{remark}{Remark}[section]

\usepackage{enumitem}
\usepackage[vlined,ruled]{algorithm2e}
\usepackage{subfig}
\usepackage{xcolor,color}
\usepackage{graphicx}
\usepackage{wrapfig}
\usepackage{mathtools}
\usepackage{multirow,multicol}
\usepackage{booktabs}
\usepackage{lipsum}
\usepackage{sidecap}
\usepackage{cleveref}
\definecolor{deepmagenta}{rgb}{0.8, 0.0, 0.8}
\usepackage{letltxmacro}
\renewcommand{\eqref}{Eq.~\ref}
\usepackage{tabularx}
\usepackage{multirow}
\usepackage{multicol}
\usepackage{thm-restate}

\definecolor{mydarkblue}{rgb}{0,0.08,0.45}
\hypersetup{
 colorlinks=true,
 citecolor=mydarkblue,
 linkcolor=mydarkblue,
 urlcolor=blue}
 
\usepackage[export]{adjustbox}

\newcommand{\opt}{\texttt{GradOptStep}}

\newcommand{\allk}{{(1:K)}}
\renewcommand{\k}{{(k)}}
\newcommand{\hG}{\widehat{G}}

\newcommand{\hvG}{\widehat{g}}
\newcommand{\oG}{\overline{G}}

\newcommand{\rvX}{X} %{\overrightarrow{X}}
\newcommand{\rvW}{W} %{\overrightarrow{W}}
\newcommand{\bw}{\overline{\vw}}

\renewcommand{\tG}{\widetilde{G}}
\newcommand{\tvG}{\widetilde{g}}
\newcommand{\te}{\widetilde{\vve}}

\newcommand{\uniond}{{d}}

\newcommand{\hpi}{\hat{\pi}}
\newcommand{\Sp}{\mathbb{S}_p}

\newcommand{\deq}{\overset{d.}{=}}
\newcommand{\tmin}{\text{\rm min}}
\newcommand{\tmax}{\text{\rm max}}

\newcommand{\gn}[1]{ { \| #1  \|_{l_1/l_2}} }
\renewcommand{\tr}{\text{\rm trace}}

\renewcommand{\tB}{B^*}
\newcommand{\tvB}{\beta^*}

\newcommand{\hB}{\widehat{B}}
\newcommand{\hZ}{\widehat{Z}}
\newcommand{\hT}{\widehat{T}}

\newcommand{\hvZ}{\widehat{\vz}}
\newcommand{\hvB}{\widehat{\beta}}
\newcommand{\Xk}{\mX^{(k)}}
\newcommand{\vW}{\vw}
\newcommand{\vB}{\beta}

\newcommand{\norm}[1]{\left\| {#1} \right\| }

\newcommand{\vU}{\vu}
\newcommand{\vY}{\vy}

\renewcommand{\k}{{(k)}}

\newcommand{\hSigma}{\widehat{\Sigma}}
\newcommand{\rsupp}{\text{RS}}

\usepackage{stmaryrd}
\usepackage[]{algorithm2e}

\SetKwComment{Comment}{$\triangleright$\ }{}

\makeatletter
\newcommand{\xMapsto}[2][]{\ext@arrow 0599{\Mapstofill@}{#1}{#2}}
\def\Mapstofill@{\arrowfill@{\Mapstochar\Relbar}\Relbar\Rightarrow}
\makeatother

\title{\Large  
Multi-task Learning of Order-Consistent Causal Graphs
}

\author{%
  Xinshi Chen\thanks{Work done partially during the visit at MBZUAI (Mohamed bin Zayed University of Artificial Intelligence)} \\
  Georgia Institute of Technology \\
 \texttt{xinshi.chen@gatech.edu}
  \And
  Haoran Sun\\
  Georgia Institute of Technology \\
  \texttt{haoransun@gatech.edu}
  \And
  Caleb Ellington \\
  Carnegie Mellon University\\
  \texttt{cellingt@cs.cmu.edu}
  \And
  Eric Xing\\
  Carnegie Mellon University\\
  MBZUAI \\
  \texttt{eric.xing@mbzuai.ac.ae}
  \And
  Le Song\\
    BioMap\\
  MBZUAI \\
  \texttt{le.song@mbzuai.ac.ae}
}

\begin{document}
\maketitle

\begin{abstract}
We consider the problem of discovering $K$ related Gaussian directed acyclic graphs (DAGs), where the involved graph structures share a consistent causal order and sparse unions of supports. Under the multi-task learning setting, we propose a $l_1/l_2$-regularized maximum likelihood estimator (MLE) for learning $K$ linear structural equation models. We theoretically show that the joint estimator, by leveraging data across related tasks, can achieve a better sample complexity for recovering the causal order (or topological order) than separate estimations. Moreover, the joint estimator is able to recover non-identifiable DAGs, by estimating them together with some identifiable DAGs.  Lastly, our analysis also shows the consistency of union support recovery of the structures. To allow practical implementation, we design a continuous optimization problem whose optimizer is the same as the joint estimator and can be approximated efficiently by an iterative algorithm. We validate the theoretical analysis and the effectiveness of the joint estimator in experiments.
\end{abstract}

\section{Introduction}

Estimating causal effects among a set of random variables is of fundamental importance in many disciplines such as genomics, epidemiology, health care and finance~\citep{friedman2000using,robins2000marginal,aguilera2011bayesian,nicholson2014applications,zhang2013integrated,peters2017elements}. Therefore, designing and understanding methods for causal discovery is of great interests in machine learning. 

Causal discovery from finite observable data is often formulated as a directed acyclic graph (DAG) estimation problem in graphical models. A major class of DAG estimation methods are score-based, which search over the space of all DAGs for the best scoring one. However, DAG estimation remains a very challenging problem from both the \textit{computational} and \textit{statistical} aspects~\citep{chickering1996learning}. On the one hand, the number of possible DAG structures grows super-exponentially in the number of random variables, whereas the number of observational sample size is normally small. On the other hand, some DAGs are \textit{non-identifiable} from observational data even with infinitely many samples. 

Fortunately, very often multiple related DAG structures need to be estimated from data, which allows us to leverage their similarity to improve the estimator. For instance, in bioinformatics, gene expression levels are often measured over patients with different subtypes~\citep{cai2016joint,liu2019joint} or under various experimental conditions~\citep{shimamura2010inferring}. 
In neuroinformatics, fMRI signals are often recorded for multiple subjects for studying the brain connectivity network~\citep{shimizu2012joint,zhang2020causal}. In these scenarios, multiple datasets will be collected, and their associated DAGs are likely to share similar characteristics. %
Intuitively, it may be beneficial to estimate these DAGs jointly.

In this paper, we focus on the analysis of the multi-task DAG estimation problem where the DAG structures can be related by a consistent causal order and a (partially) shared sparsity pattern, but allowed to have different connection strengths and edges, and differently distributed variables. In this setting, we propose a joint estimator for recovering multiple DAGs based on a group norm regularization. We prove that the joint $l_1/l_2$-penalized maximum likelihood estimator (MLE) can recover the causal order better than individual estimators. 

Intuitively, it is not surprising that joint estimation is beneficial. However, our results provide a quantitative characterization on the improvement in sample complexity and the conditions under which such improvement can hold. We show that:
\begin{itemize}[leftmargin=*,nolistsep]
    \item For identifiable DAGs, if the shared sparsity pattern (union support) size $s$ is of order $\gO(1)$ in $K$ where $K$ is the number of tasks (DAGs), then the \textit{effective} sample size for order recovery will be $nK$ where $n$ is the sample size in each problem. Furthermore, as long as $s$ is of order $o(\sqrt{K})$ in $K$, the joint estimator with group norm regularization leads to an improvement in sample complexity.
    \item A non-identifiable DAG cannot be distinguished by single-task estimators even with indefinitely many observational data. However,  non-identifiable DAGs can be recovered by our joint estimator if they are estimated together with other identifiable DAGs.
\end{itemize}
Apart from the theoretical guarantee, we design an efficient algorithm for approximating the joint estimator through a formulation of the combinatorial search problem to a continuous programming. This continuous formulation contains a novel design of a learnable masking matrix, which plays an important role in ensuring the acyclicity and shared order for estimations in all tasks. An interesting aspect of our design is that we can learn the masking matrix by differentiable search over a continuous space, but the optimum must be contained in a discrete space of cardinality $p!$ (reads $p$ factorial, where $p$ is the number of random variables).

We conduct a set of synthetic experiments to demonstrates the effectiveness of the algorithm and validates the theoretical results. Furthermore, we apply our algorithm to more realistic single-cell expression RNA sequencing data generated by SERGIO~\citep{dibaeinia2020sergio} based on real gene regulatory networks. 

The remainder of the paper is organized as follows. 
In Section~\ref{sec:background}, we introduce the linear structural equation model (SEM) interpretation of Gaussian DAGs and its properties. Section~\ref{sec:results} is devoted to the statement of our main results, with some discussion on their consequences and implications. In Section~\ref{sec:algorithm}, we present the efficient algorithm for approximating the joint estimator. Section~\ref{sec:related_work} summarizes related theoretical and practical works. Experimental validations are provided in Section~\ref{sec:experiments}.

\section{Background}
\label{sec:background}

A substantial body of work has focused on the linear SEM interpretation of Gaussian DAGs~\citep{van2013ell_,aragam2017learning,zheng2018dags,aragam2019globally}. Let $\rvX = (\rvX_1, \cdots, \rvX_p)$ be a $p$-dimensional random variable. Then a linear SEM reads
\begin{align}
\vspace{-3mm}
    \rvX = \tG_0^\top \rvX + \rvW, \quad \rvW \sim \gN(0,\Omega_0), \label{eq:lsem}
\end{align}
where $\Omega_0$ is a $p\times p$ positive diagonal matrix which indicates the variances of the noise $W$. $\tG_0\in \R^{p\times p}$ is the connection strength matrix or adjacency matrix. Each nonzero entry $\tG_{0ij}$ represents the direct causal effect of $X_i$ on $X_j$. This model implies that $X$ is Gaussian, $\rvX \sim \gN(0,\Sigma)$, where
\begin{align}
    \Sigma:= (I - \tG_0)^{-\top}\Omega_0 (I - \tG_0)^{-1}.
    \label{eq:Sigma}
    \vspace{-2mm}
\end{align}
{\bf Causal order} $\pi_0${\bf.} 
The nonzero entries of $\tG_0$ defines its causal order (also called topological order), which informs possible ``parents'' of each variable. A causal order can be represented by a permutation $\pi_0$ over $[p]:=(1,2,\cdots,p)$. We say $\tG_0$ is \textbf{consistent} with $\pi_0$ if and only if
\begin{align}
    \tG_0{}_{ij}\neq 0 \Rightarrow \pi_0(i)< \pi_0(j). \label{eq:permutation}
\end{align}
There could exist more than one permutations that are consistent with a DAG structure $\tG_0$, so we denote the set of permutations that satisfy \eqref{eq:permutation} by $\Pi_0$. Once a causal order $\pi_0$ is identified, the connection strengths $\tG_0$ can be estimated by ordinary least squares regression which is a comparatively easier problem. 

\textbf{Identifiability and equivalent class.} However, estimating the true causal order $\pi_0$ is very challenging, largely due to the existence of the {equivalent class} described below. 

Let $\sS_p$ be the set of all permutations over $[p]$. For every  $\pi\in \sS_p$, there exists a connection strength matrix $\tG(\pi)$, which is consistent with $\pi$, and a diagonal matrix
$    \Omega(\pi)=\text{diag}\sbr{\sigma_1(\pi)^2,\cdots, \sigma_p(\pi)^2}
$ such that the variance $\Sigma$ in \eqref{eq:Sigma} equals to 
\begin{align}
    \Sigma = (I - \tG(\pi))^{-\top}\Omega(\pi) (I - \tG(\pi))^{-1}, \label{eq:G-pi}
\end{align}
and therefore the random variable $\rvX$ in \eqref{eq:lsem} can be equivalently described by the model \citep{aragam2017learning}:
\begin{align}
    \rvX = \tG(\pi)^\top \rvX + \rvW(\pi),\quad \rvW(\pi) \sim \gN(0,\Omega(\pi)).
\end{align}
Without further assumption, the true underlying DAG, $\tG_0=\tG(\pi_0)$, \textbf{cannot} be identified from the 
\begin{equation}
    \text{equivalent class: }\quad \{\tG(\pi): \pi \in \sS_p\}\label{eq:equivalent}
\end{equation}
based on the distribution of $\rvX$, even if infinitely many samples are observed.

\section{Joint estimation of multiple DAGs}

\label{sec:results}
% \vspace{-2mm}

In the multi-task setting, we consider $K$ linear SEMs: 
\begin{align*}
    \rvX^\k = \tG_0^\k{}^\top \rvX^\k + \rvW^\k,\quad \text{for }k=1,\cdots,K.
\end{align*}
The superscript notation ${}^\k$ indicates the $k$-th model.
As mentioned in Sec~\ref{sec:background}, each model defines a random variable $\rvX^\k\sim\gN(0,\Sigma^\k)$ with $\Sigma^\k = (1-\tG_0^\k)^{-\top}\Omega_0^\k (1-\tG_0^\k)^{-1}$.

\subsection{Assumption}

\textbf{Similarity.} In Condition~\ref{cond:order} and Definition~\ref{def:union}, we make a few assumptions about the similarity of these models. Condition~\ref{cond:order} ensures the causal orders in the $K$ DAGs do not have conflicts. %, so that it is sensible to jointly recover a single order for all the $K$ DAGs. 
Definition~\ref{def:union} measures the similarity of the sparsity patterns in the $K$ DAGs. If the sparsity patterns are strongly overlapped, then the size of the support union $s$ will be small. We do not enforce a strict constraint on the support union, but we will see in the later theorem that a smaller $s$ can lead to a better recovery performance.
\begin{condition}[Consistent causal orders]
    \label{cond:order}
    There exists a nonempty set of permutations $\Pi_0\subseteq \sS_p$ such that $\forall \pi_0\in\Pi_0$, it holds for all $k\in[K]$ that $\tG_0{}_{ij}^\k\neq 0 \Rightarrow \pi_0(i)< \pi_0(j)$. 
\end{condition} 
\begin{definition}[Support union]
    \label{def:union} Recall $\tG(\pi)$ in \eqref{eq:equivalent}. The support union of the $K$ DAGs associated with permutation $\pi$ is denoted by
    $S(\pi):= \big\{(i,j):\exists k\in[K]\, s.t.\, \tG^\k_{ij}(\pi)\neq 0\big\}$. The support union of the $K$ true DAGs $\tG_0^\k$ is $S_0:= S(\pi_0)$. We further denote $s_0:= \abs{S_0}$ and $s:=\sup_{\pi\in\sS_p}\abs{S(\pi)}$. %\Le{Suppose $s_0$ is small, is there any discussion on how large $s$ can be?}
\end{definition}

\textbf{Identifiability.} To ensure the consistency of the estimator based on least squared loss, we first assume that the DAGs to be recovered are minimum-trace DAGs.
\begin{condition}[Minimum-trace]
\label{cond:min-trace}
Recall the equivalent class defined in \eqref{eq:G-pi} to \eqref{eq:G-pi}. Assume for all $k=1,\cdots,K$, $\tr(\Omega_0^{(k)}) = \min_{\pi\in\sS_p}\tr(\Omega(\pi)^{(k)})$. 
\end{condition}
However, the minimum-trace DAG may not be unique without further assumptions, making the true DAG indistinguishable from other minimum-trace DAGs. Therefore, we consider the equal variance condition in Condition~\ref{cond:equal-variance} which ensures the uniqueness of the minimum-trace DAG. In this paper, we assume the first $K'\leq K$ models satisfy this condition, so they are identifiable. We do not make such an assumption on the other $K-K'$ models, so the $K-K'$ models may not be identifiable.  
\begin{condition}[Equal variance]
    \label{cond:equal-variance} For all $k=1,\cdots,K'$ with $K'\leq K$, the noise 
    $
        W^\k \sim \gN(0,\Omega_0^\k)$ has equal variance with $\Omega_0^\k = \sigma_0^{(k)} I_p
    $.
\end{condition}

\vspace{-1mm}
\subsection{$l_1/l_2$-penalized joint estimator }
% \vspace{-2mm}
\vspace{-1mm}

Denote the sample matrix by $\mX^\k$ whose row vectors are $n$ i.i.d. samples from $\gN(0,\Sigma^\k)$. 
Based on the task similarity assumptions, we propose the following {$l_1/l_2$-penalized joint maximum likelihood estimator} (MLE) for jointly estimating the connection strength matrices $\{\tG{}^\k_0\}_{k=1}^K$:
\begin{align}
    \hpi, \{\hG^\k\} = \argmin_{\pi\in\Sp,\,\cbr{G^\k \in \sD(\pi)} }  \sum_{k=1}^K \frac{1}{2n}\|\mX^\k - \mX^\k G^\k(\pi)\|_F^2 + \lambda \gn{G^\allk(\pi) }. \label{eq:joint-estimator}
\end{align}
Similar to the notation in \eqref{eq:G-pi}, $G^\k(\pi)$ indicates its consistency with $\pi$. $\sD(\pi)$ denote the space of all DAGs that are consistent with $\pi$. It is notable that a single $\pi$ shared across all $K$ tasks is optimized in \eqref{eq:joint-estimator}, which respects Condition~\ref{cond:order}. The group norm over the set of $K$ matrices is defined as 
\begin{align*}
   \textstyle \gn{G^\allk(\pi)}:= \sum_{i=1}^p \sum_{j=1}^p \|G^\allk_{ij}\|_2,\quad \text{where }G^\allk_{ij}:=[G^{(1)}_{ij}(\pi),\cdots, G^{(K)}_{ij}(\pi)].
\end{align*}
It will penalize the size of union support in a soft way. When $K=1$, this joint estimator will be reduced to the $l_1$-penalized maximum likelihood estimation.

\begin{remark}
The optimization in \eqref{eq:joint-estimator} is used for analysis only. A continuous program with the same optimizer will be discussed in Section~\ref{sec:algorithm} for practical implementation. 
\end{remark}

\subsection{Main result: causal order recovery}

We start with a few pieces of notations and definitions. Then the theorem statement follows.
\begin{definition}
\label{def:1}
Let $\tvG_j^\k(\pi)$ denote the $j$-th column of $\tG^\k(\pi)$. Let
\begin{align*}
   \textstyle d:=  \sup_{j\in[p],\pi\in\mathbb{S}_p}\big|\cup_{k\in[K]} \text{supp}(\tvG^\k_j(\pi))\big|,\quad g_\tmax:=  \sup_{\pi\in\sS_p, (i,j)\in S(\pi)} \|\tG^\allk_{ij}(\pi) \|_2 / \sqrt{K}.
\end{align*}
\end{definition}
In Definition~\ref{def:1}, $d$ is the maximal number of parents (in union) in the DAGs $\tG^\k(\pi)$, which is also a measure of the sparsity level. $g_\tmax$ is bounded by the maximal entry value in the matrices $\tG^\k(\pi)$.
\begin{condition}[Bounded spectrum]
\label{cond:spec}
Assume for all $k=1,\cdots,K$, the covariance matrix $\Sigma^\k$ is positive definite. 
There exists constants $0<\Lambda_\tmin \leq \Lambda_\tmax < \infty $ such that for all $k=1,\cdots,K$,
\begin{enumerate}[leftmargin=0.8cm,nolistsep]
    \item[(a)] all eigenvalues of $\Sigma^\k$ are upper bounded by $\Lambda_\tmax$;
    \item[(b)] all eigenvalues of $\Sigma^\k$ are lower bounded by $\Lambda_\tmin$.
\end{enumerate}
\end{condition}
\vspace{0.1cm}
\begin{condition}[Omega-min]
\label{cond:iden}
There exists a constant $\eta_w > 0$ such that for any permutations $\pi\notin \Pi_0$,
\begin{align*}
    \textstyle \frac{1}{pK'} \sum_{k=1}^{K'}\sum_{j=1}^p (\sigma_j^\k(\pi)^2 - \sigma_{0}^{(k)}{}^2)^2 > \frac{1}{\eta_w}.
\end{align*}
\end{condition}

Condition \ref{cond:iden} with $K'=K=1$ is called `omega-min' condition in \citep{van2013ell_}, so we follows this terminology. In some sense, when $\eta_w$ is larger, $\sigma_j(\pi)$ with $\pi\notin\Pi_0$ is allowed to deviate less from the true variance $\sigma_j(\pi_0)=\sigma_0$ with $\pi_0\in\Pi_0$, which will make it more difficult to separate the set $\Pi_0$ from its~complement in a finite sample scenario. 
Ideally, we should allow $\eta_w$ to be large, so that the recovery is not only restricted to easy problems.

Now we are ready to present the recovery guarantee for causal order.  Theorem~\ref{thm:order-recovery} is a specific statement when the regularization parameter $\lambda$ follows the classic choice in (group) Lasso problems. A more general statement which allows other $\lambda$ is given in Appendix~\ref{app:thm1} along with the proof.

\begin{theorem}[Causal order recovery]
\label{thm:order-recovery}
Suppose we solve the joint optimization in \eqref{eq:joint-estimator} with specified regularization parameter $\lambda = \sqrt{\frac{p\log p}{n}}$ for a set of $K$ problems that satisfy Condition~\ref{cond:order}, \ref{cond:spec} (a), \ref{cond:min-trace}, \ref{cond:equal-variance} and \ref{cond:iden}. If the following conditions are satisfied:
\vspace{-2mm}
    \begin{align}
      \theta(n,K,K{}',p,s) & := \frac{p}{s}\sqrt{\frac{n}{p\log p}\frac{K'{}^2}{K}} >  \kappa_1 \eta_w, \label{eq:sample-complexity}\\
       n & \geq c_1 \log K + c_2 (d+1)\log p, \label{eq:n-cond} \\
       K & \leq \kappa_2 p\log p,  \label{eq:k-cond}
    \end{align}
then the following statements hold true:
\begin{enumerate}[leftmargin=0.8cm]
    \item[(a)] With probability at least $
        1 - c_3\exp\rbr{-\kappa_4 (d+1)\log p} - \exp\rbr{-c_4 n}$, it holds that
    \begin{align*}
        \hpi \in \Pi_0.
    \end{align*}
    % where $\hat{s}:=\big|\{(i,j):\exists k \,s.t.\, \hG^\k_{ij}\neq 0\}\big| $ is the sparsity of the estimated support union.
    \item[(b)]If in addition, $n$ satisfies $n\geq \kappa_5 \hat{d}(\log K + \log p)$ with $\hat{d}:=\max_{j\in[p],k\in[K]}\|\hvG_j^\k - \tvG_j^\k\|_0$, and Condition~\ref{cond:spec} (b) holds, then with probability at least 
   $ 1 - c_3\exp\rbr{-\kappa_4 (d+1)\log p} - \exp\rbr{-c_4 n}-\exp(-\log p- \log K)$,
    \begin{align*}
           \textstyle\frac{1}{K}\sum_{k=1}^K \|\hG^\k - \tG_0^\k\|_F^2  = \gO\rbr{  s_0 \sqrt{\frac{p\log p}{nK}} }.
   \end{align*}
\end{enumerate}
In this statement, $c_1,c_2,c_3,c_4$ are universal constants (i.e., independent of $n,p,K,s,\Sigma^\k$), $\kappa_1,\kappa_2,\kappa_3,\kappa_4$ are constants depending on $g_\tmax$ and $\Lambda_\tmax$, and $\kappa_5$ is a constant depending on $\Lambda_\tmin$.
\end{theorem}

\textbf{Discussion on \eqref{eq:sample-complexity}-\eqref{eq:k-cond}.} (i) In \eqref{eq:sample-complexity}, Theorem~\ref{thm:order-recovery} identifies a \textit{sample complexity parameter} $\theta(n,K,K',p,s)$ . Following the terminology in \citep{obozinski2011support}, our use of the term ``sample complexity'' for $\theta$ reflects the dominant role it plays in our analysis as the rate at which the sample size much grow in order to obtain a consistent causal order. More precisely, for scalings $(n,K,K',p,s)$ such that $\theta(n,K,K',p,s)$ exceeds a fixed critical threshold $\kappa_1 \eta_w$, we show that the causal order can be correctly recovered with high probability. 

(ii) The additional condition for the sample size $n$ in \eqref{eq:n-cond} is the sample requirement for an ordinary linear regression problem. It is in general much weaker than \eqref{eq:sample-complexity}, unless $K$ grows very large.

(iii) The last condition in \eqref{eq:k-cond} on $K$ could be relaxed if a tighter analysis on the distribution properties of Chi-squared distribution is available. However, it is notable that this restriction on the size of $K$ has already been weaker than many related works on multi-task $\ell_1$ sparse recovery, which either implicitly treat $K$ as a constant~\citep{obozinski2011support,wang2014collective} or assume $K=o(\log p)$~\citep{wang2020high,lee2020bayesian}.

\textbf{Recovering non-identifiable DAGs.} A direct consequence of Theorem~\ref{thm:order-recovery} is that as long as the number of identifiable DAGs $K'$ is non-zero, the joint estimator can recover the causal order of non-identifiable DAGs with high probability. This is not achievable in separate estimation even with infinitely many samples. Therefore, we show that how the information of identifiable DAGs helps recover the non-identifiable ones. 

\textbf{Effective sample size.} As indicated by $\theta(n,K,K',p,s)$ in \eqref{eq:sample-complexity}, the effective sample size for recovering the correct causal order is $\frac{nK'{}^2}{K}$ if the support union size $s$ is of order $\gO(1)$ in $K$. To show the improvement in sample complexity, it is more fair to consider the scenario when the DAGs are identifiable, i.e., $K'=K$. In this case, it is clear that the parameter $\theta(n,K,K,p,s)$ indicates a lower sample complexity relative to separate estimation 
as long as $s$ is of order $o(\sqrt{K})$. 

\textbf{Separate estimation.} Consider the special case of a single task estimation with $K=K'=1$, in which the joint estimator reduces to $\ell_1$-penalized MLE. We discuss how our result can recover previously known results for single DAG recovery. Unfortunately, existing analyses  were conducted under different frameworks with different conditions. \citep{van2013ell_} and \citep{champion2018inferring} are the most comparable ones since they were based on the same omega-min condition (i.e., Condition~\ref{cond:iden}), but they chose a smaller regularization parameter $\lambda$. In our proof, the sample complexity parameter is derived from  $\theta(n,K,K',p,s)= pK'/\rbr{s\lambda \sqrt{K}}$ and \eqref{eq:sample-complexity} is $pK'/\rbr{s\lambda \sqrt{K}} > \kappa_1 \eta_w$. When $K=K'=1$, this condition matches what is identified in \citep{van2013ell_} and \citep{champion2018inferring} for recovering the order of a single DAG.

\textbf{Error of $\hG^\k$.} To compare the estimation of $\hG^\k$ to the true DAG $\tG_0^\k$, Theorem~\ref{thm:order-recovery}~(b) says the averaged error in F-norm goes to zero when $nK\rightarrow \infty$. It decreases in $K$ as long as $s=o(K)$.

To summarize, Theorem~\ref{thm:order-recovery} analyzes the causal order consistency for the joint estimator in \eqref{eq:joint-estimator}. Order recovery is the most challenging component in DAG estimation. After $\hpi\in\Pi_0$ has been identified, the DAG estimation becomes $p$ linear regression problems that can be solved separately. Theorem~\ref{thm:order-recovery} (b) only shows the  estimation error of connection matrices in F-norm. To characterize the structure error, additional conditions are required. 

\subsection{Support union recovery}

Theorem~\ref{thm:order-recovery} has shown that $\hpi=\pi_0\in\Pi_0$ holds with high probability. Consequently, the support recovery analysis in this section is conditioned on this event. In fact, given the true order $\pi_0$, what remains is a set of $p$ separate $l_1/l_2$-penalized group Lasso problems, in each of which the order $\pi_0$ plays a role of constraining the support set by the set $S_j(\pi_0):=\{i: \pi_0(i) < j\}$. 
However, we need to solve $p$ such problems simultaneously where $p$ is large. A careful analysis is required, and directly combining existing results will not give a high recovery probability.

In the following, we impose a set of conditions and definitions, which are standard in many $l_1$ sparse recovery analyses~\citep{obozinski2011support,wang2014collective}, after which theorem statement follows. See Appendix~\ref{app:thm2} for the proof.
\begin{definition} \label{def:thm2}
The union support of $j$-th columns of $\{\tG^\k_0\}_{k\in[K]}$ is denoted by $
    \rsupp_j:=\{i\in[p]: \exists k\in[K]\,s.t.\,\tG^\k_{0ij}\neq 0 \}
$. The maximal cardinality is $r_\tmax:= \sup_{j\in[p]}\big|\rsupp_j\big|$.
\end{definition}

\begin{definition} \label{def:thm2-2}
$
    \rho_u: = \sup_{k\in[K],j\in[p],S=\rsupp_j}\max_{i\in S^c} \big(\Sigma_{S^c S^c | S}^\k\big)_{ii}
$ is the maximal diagonal entry of the conditional covariance matrices, where $\Sigma_{S^c S^c | S}^\k:=\Sigma^\k_{S^c S^c} - \Sigma^\k_{S^c S}(\Sigma^\k_{SS})^{-1}\Sigma_{SS^c}^\k$.
\end{definition}

\begin{definition} \label{def:thm2-3}
$g_\tmin:= \inf_{(i,j)\in S_0} \|\tG_{0ij}^\allk\|_2/\sqrt{K}$ represents the signal strength.
\end{definition}

\begin{condition}[Irrepresentable condition]
\label{cond:irrepresentable}
There exists a fixed parameter $\gamma\in(0,1]$ such that
    $\sup_{j\in[p],S=\rsupp(\tvG^\allk_{0j})}\norm{A(S) }_\infty \leq 1-\gamma $, where $A(S)_{ij}:=\sup_{k\in[K]}\big|\big(\Sigma_{S^cS}^\k(\Sigma_{SS}^\k)^{-1}\big)_{ij} \big|$.
\end{condition}

\begin{theorem}[Union support recovery]
\label{thm:support}

Assume on the subset of probability space where $\{\hpi \in \Pi_0\}$ holds, and assume Condition~\ref{cond:irrepresentable}. Assume the following conditions are satisfied
    \begin{align}
       n &\geq \kappa_6 r_\tmax \log p,  \label{eq:cond-n-thm2}\\
        K &\leq c_7 \log p, 
        \label{eq:cond-k-thm2} \\
        \sqrt{ \frac{8\Lambda_\tmax \log p}{\Lambda_\tmin n} } &+ \frac{2 }{\Lambda_\tmin}\sqrt{\frac{p\log p}{n}{\frac{r_\tmax}{K}}} = o(g_\tmin),
        \label{eq:complexity-thm2}
    \end{align}
    where $\kappa_6$ is a constant depending on $\gamma,\Lambda_\tmin,\rho_u,\sigma_\tmax$, and $c_7$ is some universal constant. Then w.p. at least
    $
        1-  r_\tmax \exp\rbr{-c_5 K\log p}- \exp\rbr{-c_6 \log p} - c_8K\exp\rbr{-c_9(n-r_\tmax-\log p)}
    $,
    the support union of $\hG^\allk$ is the same as that of $\tG_0^\allk$, and that
    $
        \big\|\hG^\allk - \tG^\allk\big\|_{l_\infty/l_2}/\sqrt{K} = o(g_\tmin)
    $.
\end{theorem}

\textbf{Discussion on \eqref{eq:cond-n-thm2} and \eqref{eq:cond-k-thm2}.} \textit{(i)} \eqref{eq:cond-n-thm2} poses a sample size condition. The value $r_\tmax$ is the sparsity overlap defined in Definition~\ref{def:thm2} (i). It takes value in the interval [$d,\min\{s_0,p,dK\}$], depending on the similarity in sparsity pattern.

\textit{(ii)} The restriction on $K$ in \eqref{eq:cond-k-thm2} plays a similar role as \eqref{eq:k-cond} in Theorem~\ref{thm:order-recovery}. This a stronger restriction, but also guarantees the stronger result of support recovery. Existing analyses on $l_1/l_2$-penalized group Lasso were not able to relax this constraint, neither, so some of them treated $K$ as a constant in the analysis~\citep{obozinski2011support,wang2014collective}. Recall that in Theorem~\ref{thm:order-recovery}, we were able to allow $K=\gO(p\log p)$. Technically, this was achieved because in the proof of Theorem~\ref{thm:order-recovery}, we avoid analyzing the general recovery of group Lasso, but only its null-consistency (i.e., the special case of true structures having zero support), where tighter bound can be derived and it is sufficient for order recovery.

\textbf{Benefit of joint estimation.}
\eqref{eq:complexity-thm2} plays a similar role as \eqref{eq:sample-complexity} in Theorem~\ref{thm:order-recovery}. It specifies a rate at which the sample size must grow for successful union support recovery. As long as $r_\tmax$ is of order $o(\sqrt{K})$, $K$ will effectively reduce the second term in \eqref{eq:sample-complexity}. Apart from that, the recover probability specified in Theorem~\ref{thm:support} grows in $K$. 

\section{Algorithm}
\label{sec:algorithm}

Solving the optimization in \eqref{eq:joint-estimator} by searching over all permutations $\pi\in\Sp$ is intractable due to the large combinatorial space. Inspired by the smooth
characterization of acyclic graph~\citep{zheng2018dags}, we propose a continuous optimization problem, whose optimizer is the same as the estimator in \eqref{eq:joint-estimator}. Furthermore, we will design an efficient iterative algorithm to approximate the solution.

\subsection{Continuous program}

We convert \eqref{eq:joint-estimator} to the following constrained continuous program
\begin{align}
    \textstyle \min_{\substack{T\in \R^{p\times p}  \\ G^{(1)},\cdots,G^{(K)}\in \R^{p\times p}}}  & \sum_{k=1}^K \frac{1}{2n} \left\|\mX^\k - \mX^\k \oG{}^\k \right\|_F^2 + \lambda \gn{\oG{}^\allk } + \rho \|\vone_{p\times p} - T\|_F^2 \label{eq:cont-opt} \\
    & \text{subject to}  \quad  h(T):=\tr(e^{T\circ T})-p =0, \label{eq:dag-const}
\end{align}
where $\oG{}^\k := G^\k \circ T$ is element-wise multiplication between $G^\k$ and $T$, and $\vone_{p\times p}$ is a $p\times p$ matrix with entries equal to one. \eqref{eq:dag-const} is a smooth `DAG-ness' constraint proposed by NOTEARS~\citep{zheng2018dags}, which ensures $T$ is acyclic. One can also use $h(T):=\tr((I + T \circ T)^p)-p$ proposed in~\citep{yu2019dag}.

\begin{wrapfigure}[8]{R}{0.5\textwidth}
\vspace{-3mm}
\begin{minipage}[t]{.25\textwidth}\vspace*{0pt}
\begin{gather*}
    \pi = (1,2,3,4)\\
    \rotatebox[origin=c]{90}{$\Leftrightarrow$}\\
    T=\sbr{
    \begin{array}{cccc}
       0  & 0 & 0 & 0  \\
        1 & 0 & 0 & 0 \\
        1 & 1& 0 & 0 \\
        1 & 1& 1 & 0
    \end{array}}
\end{gather*}
\end{minipage}\hfill%
\begin{minipage}[t]{.25\textwidth}
\begin{gather*}
    \pi = (4,2,1,3)\\
    \rotatebox[origin=c]{90}{$\Leftrightarrow$}\\
    T=\sbr{
    \begin{array}{cccc}
        0 & 0 & 1 & 0 \\
        1 & 0 & 1 & 0 \\
        0 & 0 & 0 & 0 \\
        1 & 1 & 1 & 0
    \end{array}}
\end{gather*}
\end{minipage}
\end{wrapfigure}
We would like to highlight the  novel and interesting design of the matrix $T$ in \eqref{eq:cont-opt}. What makes \eqref{eq:joint-estimator} difficult to solve is the requirement that $\{G^\k\}$ must be DAGs and share the same order. A straightforward idea is to apply the smooth acyclic constraint to every $G^\k$, but it is not clear how to enforce their consistent topological order. Our formulation realizes this by a single matrix $T$.

To better understand the design rationale of $T$, recall in \eqref{eq:permutation} that a matrix $G$ is a DAG of order $\pi$ \textit{if and only if} its support set is in $\{(i,j):\pi(i)<\pi(j)\}$. The matrix $T$ plays a role of restricting the support set of $G^\k$ by masking its entries. Two examples are shown above. However, unlike the learning of masks in other papers which allows $T$ to have any combinations of nonzero entries, here we need $T$ to exactly represent the support set for each $\pi$. That is $T$ is from a space $\gT_p$ with $p!$ elements: $\gT_p:=\{T\in\{0,1\}^{p\times p}: T_{ij}=1\Leftrightarrow \pi(i)<\pi(j)\}$.
% \begin{align*}
%     \gT_p:=\{T\in\{0,1\}^{p\times p}: T_{ij}=1\Leftrightarrow \pi(i)<\pi(j)\}.
% \end{align*}
Now a key question arises: \textit{How to perform a continuous and differentiable search over $\gT_p$?} The following finding motivates our design:
\begin{align*}
    T\in \gT_p \Longleftrightarrow T\in {\textstyle\argmin_{T\in\R^{p\times p}}} \cbr{ \|\vone_{p\times p} - T\|_F^2\text{ subject to } h(T)=0}. 
\end{align*}
In other words, $T$ is a continuous projection of $\vone_{p\times p}$ to the space of DAGs. We can then optimize the mask $T$ in the continuous space $\R^{p\times p}$ but the optimal solution must be an element in the discrete space $\gT_p$. This observation also naturally leads to the design of \eqref{eq:cont-opt}.   

Finally, we want to emphasize that it is important for the optimal $T$ to have binary entries. Without this property, any nonzero value $c$ can scale the $(G,T)$ pair to give an equivalent masked DAG, i.e., $G\circ T = (c G) \circ (\frac{1}{c}T)$. This scaling equivalence will make the optimization hard to solve in practice.

Proofs for the above arguments and  the equivalence between \eqref{eq:joint-estimator} and \eqref{eq:cont-opt} are in Appendix~\ref{app:continuous}.

\begin{algorithm}[t]
\SetKwInOut{Hyper}{Hyperparameters}
 \Hyper{$\rho,\alpha,\lambda,t,\delta$}
 Initialize $G^\allk, T$ randomly\;
 \For{$itr=1,\cdots,M$}{
  \For{$itr'=1,\cdots,M'$}{
    $[G^\allk, T] \gets  \opt \big( f; G^\allk, T,\beta\big)$\Comment*[r]{Gradient-based update on $f$}
    $\forall i,j\in[p],\, G^\allk_{ij} \gets \frac{G^\allk_{ij}}{\|G^\allk_{ij}\|_2} \max \cbr{ 0, \|G^\allk_{ij}\|_2 - t \lambda |T_{ij}|}$\Comment*[r]{Proximal step}
    $\forall i,j\in[p],\,T_{ij} \gets \sign(T_{ij}) \max\big\{0, |T_{ij}|-t \lambda \|G^\allk_{ij}\|_2\big\}$\Comment*[r]{Proximal step} %\Le{Don't quite understand where this update on T is coming from.}
  }
  $\beta\gets \beta + \tau h(T)$\Comment*[r]{Dual ascent}
  $\alpha\gets \alpha\cdot (1+\delta)$\Comment*[r]{Typical rule~\citep{yu2019dag}}
  }
\caption{Joint Estimation Algorithm}
\label{algo}
\vspace{-1mm}
\end{algorithm}

\subsection{Iterative algorithm}

We derive an efficient iterative algorithm using the {Lagrangian method with quadratic penalty}, which converts \eqref{eq:cont-opt} to an unconstrained problem:
\begin{gather*}
    \min_{T,G^{(1)},\cdots,G^{(K)}\in \R^{p\times p}} \max_{\beta \geq 0} \,\,\gL(G^\allk,T;\,\beta) :=   f(G^\allk,T;\,\beta) +  \lambda \gn{\oG{}^\allk },\\
     \text{where }f(G^\allk,T;\,\beta):=\sum_{k=1}^K \frac{1}{2n} \left\|\mX^\k - \mX^\k \oG{}^\k \right\|_F^2  + \rho \|\vone_{p\times p} - T\|_F^2 + \beta h(T) + \alpha h(T)^2 ,
\end{gather*}
% where $f(G^\allk,T;\,\beta):=\sum_{k=1}^K \frac{1}{2n} \left\|\mX^\k - \mX^\k \oG{}^\k \right\|_F^2  + \rho \|\vone_{p\times p} - T\|_F^2 + \beta h(T) + \alpha h(T)^2$, 
$\beta$ is dual variable, $\alpha$ is the coefficient for quadratic penalty, and $f$ is the smooth term in the objective.

We can solve this min-max problem by alternating primal updates on $(G^\allk, T)$ and dual updates on $\beta$. Due to the non-smoothness of group norm, the primal update is based on proximal-gradient method, where the
\textbf{proximal-operator} with respect to $\gn{\cdot}$ has a closed form:
\begin{gather*}
    \textstyle\big[ \argmin_{Z^\allk \in \R^{K\times p\times p}} \frac{1}{2}\sum_{k=1}^K\|Z^\k- \mX^\k\|_F^2 + c \gn{Z^\allk} \big]_{ij}^\allk \nonumber\\
    \textstyle=\frac{\mX^\allk_{ij}}{\|\mX^\allk_{ij}\|_2} \max\{0, \|\mX_{ij}^\allk\|_2-c\},
\end{gather*}
which is a group-wise soft-threshold. Since $G^\k$ and $T$ are multiplied together element-wisely inside the group norm, the proximal operator will be applied to both of them separately. 
Together with the dual update for $\beta$, 
$
    \beta \gets \beta +  \tau h(T)
$
with $\tau$ as the step size, we summarize the overall algorithm for solving \eqref{eq:cont-opt} in Algorithm~\ref{algo}.

\section{Related work}
\label{sec:related_work}

\textbf{Single DAG estimation.}  Unlike the large literature of research on {undirected} graphical models~\citep{cox1996multivariate,meinshausen2006high,yuan2007model,ravikumar2008model}, statistical guarantees for score-based DAG estimator have been available only in recent years. \citep{van2013ell_,aragam2017learning,aragam2019globally} have shown the DAG estimation consistency in high-dimensions, but they do not consider joint estimation. Nevertheless, some techniques in \citep{van2013ell_,aragam2017learning} are useful for our derivations. %Furthermore, our results with $K=1$ could match the rate in \citep{van2013ell_} and \citep{champion2018inferring}.

\textbf{Multi-task learning.} 
\textit{(i) Undirected graph estimation.} There have been extensive studies on the joint estimation of multiple undirected Gaussian graphical models~\citep{song2009keller,honorio2010multi,guo2011joint,cai2011constrained,oyen2012leveraging,danaher2014joint,kolar2010estimating,mohan2014node,peterson2015bayesian,yang2015fused,gonccalves2016multi,varici2021learning}. 
\textit{(ii) DAG estimation.} In contrast, not much theoretical work has been done for joint DAG learning. A few pieces of recent works addressed certain statistical aspects of multi-task DAG estimation \citep{liu2019joint,wang2020high,lee2020bayesian}, but \citep{wang2020high,lee2020bayesian} tackle the fewer task regime with $K=o(\log p)$, and \citep{liu2019joint} assumes the causal order is given. Another related work that we notice after the paper submission is \cite{ghoshal2019direct}, which also assumes the DAGs have a consistent causal order, but it focuses on estimating the difference between two DAGs. \textit{(iii) Linear regression.} Multi-task linear regression is also a related topic~\citep{bach2008consistency,negahban2008joint,zhao2009composite,lounici2009taking,huang2010benefit,chen2011integrating,tian2016forward}, because the techniques for analyzing group Lasso are used in our analysis~\citep{obozinski2011support,wang2014collective}.

\textbf{Practical Algorithms.} Works on practical algorithm design for efficiently solving the score-based optimization are actively conducted \citep{scanagatta2019survey,ng2019masked,Zhu2020Causal,manzour2021integer}. Our algorithm is most related to recent methods exploiting a smooth
characterization of acyclicity, including NOTEARS \citep{zheng2018dags} and several subsequent works \citep{yu2019dag,Lachapelle2020Gradient,zheng2020learning,pamfil2020dynotears}, but they only apply for single-task DAG estimation. Although algorithms for the multi-task counterpart were proposed a decade ago \citep{shimizu2012joint,tillman2011learning,tillman2009structure,xu2014pooling}, none of them leverage recent advances in characterizing DAGs and providing theoretically guarantees.

\section{Experiments}
\label{sec:experiments}

\subsection{Synthetic data}
\label{sec:syn-experiments}

The set of experiments is designed to reveal the effective sample size predicted by Theorem~\ref{thm:order-recovery}, and demonstrate the effectiveness of the proposed algorithm. In the simulations, we randomly sample a causal order $\pi$ and a union support set $S_0$. Then we randomly generate multiple DAGs that follow the order $\pi$ and have edges contained in the set $S_0$. For each DAG, we construct a linear SEM with standard Gaussian noise, and sample $n$ data points from it. On tasks with different combinations of $(p,n,s,K)$, we exam the behavior of the joint estimator, estimated by Algorithm~\ref{algo}, 
on 64 tasks and report the statistics in the following for evaluation. In this experiment, we take $K'=K$ so that all the DAGs are identifiable. This simpler case will make it easier to verify the proposed algorithm and the rates in the theorem.

\begin{wrapfigure}[11]{R}{0.35\textwidth}
    \centering
    thm:supportce{-4mm}
    \includegraphics[width=0.35\textwidth]{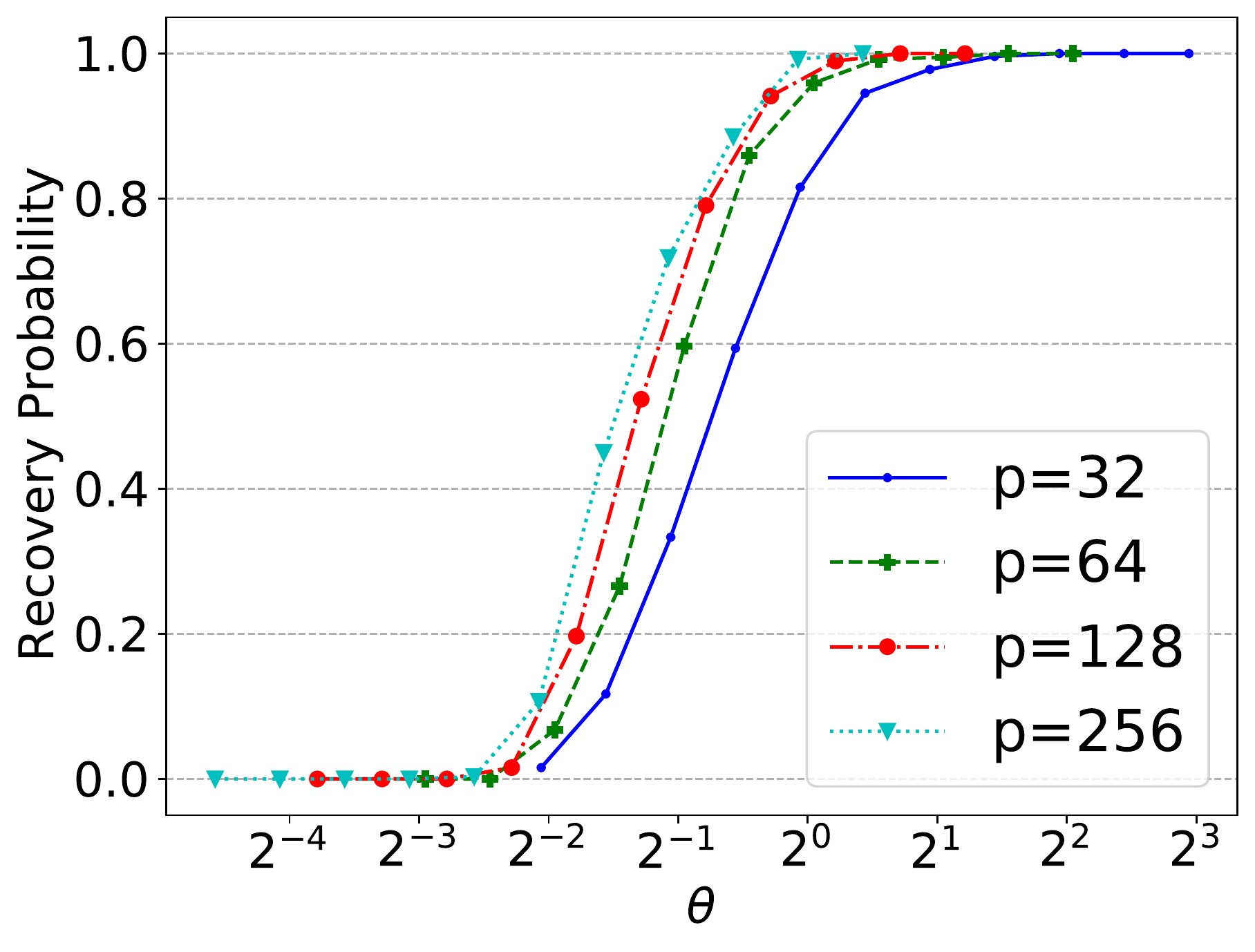}
    \vspace{-6mm}
    \caption{\small Success probability vs $\log \theta$.}
    \label{fig:transition}
\end{wrapfigure}
\textbf{Success probability for order recovery.}
For each fixed tuple $(p,n,s,K)$, we measure the sample complexity in terms of the parameter $\theta$ specified by Theorem~\ref{thm:order-recovery}. Fig~\ref{fig:transition} plots the success probability $\Pr[\hpi\in\Pi_0]$, versus $\theta=p/s\sqrt{nK/(p\log p)}$ at a logarithmic scale. Theorem~\ref{thm:order-recovery} predicts that the success probability should transition to 1 once $\theta$ exceeds a critical threshold. Curves in Fig~\ref{fig:transition} actually have sharp transitions, showing step-function behavior. The sharpness is moderated by the logarithmic scale in $x$-axis. Moreover, by scaling the sample size $n$ using $\theta$, the curves align well as predicted by the theory and have a similar transition point, even though they correspond to very different model dimensions $p$. 

\begin{figure}[h]
    \centering
    \includegraphics[width=\textwidth]{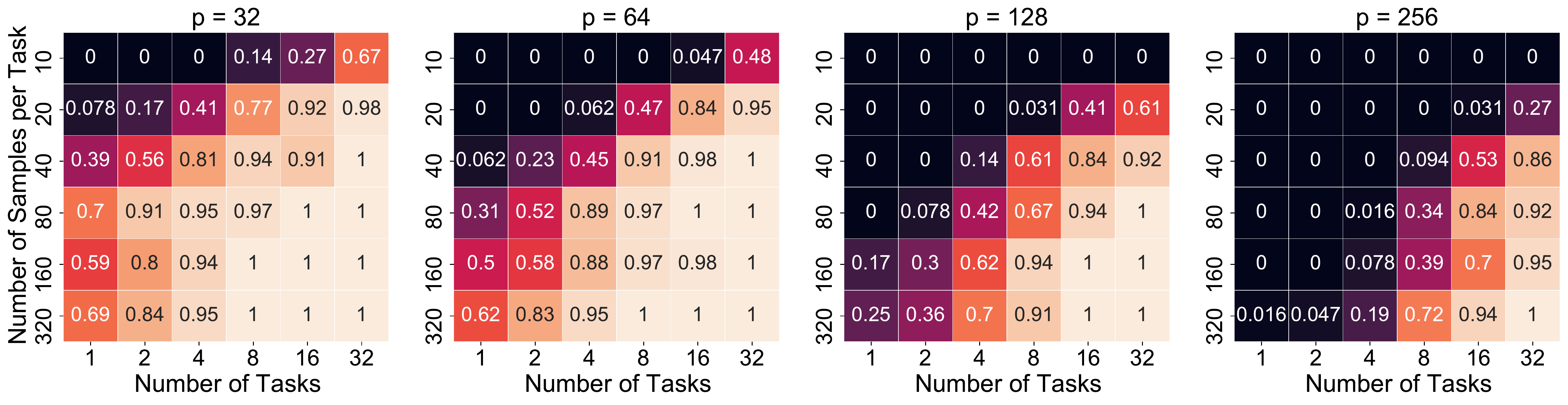}
    \vspace{-5mm}
    \caption{Heat map: Darker colors indicate lower success probability, and lighter colors  are higher.}
    \label{fig:hea_map}
\end{figure}

Fig~\ref{fig:hea_map} shows the success probability in the form of Heat Maps, where the rows indicate an increase in per task sample size $n$, and the columns indicate an increase in the number of tasks $K$. The results show that the increases in these two quantities have similar effect to the success probability.

\begin{figure}[h]
    \centering
    \includegraphics[width=0.9\textwidth]{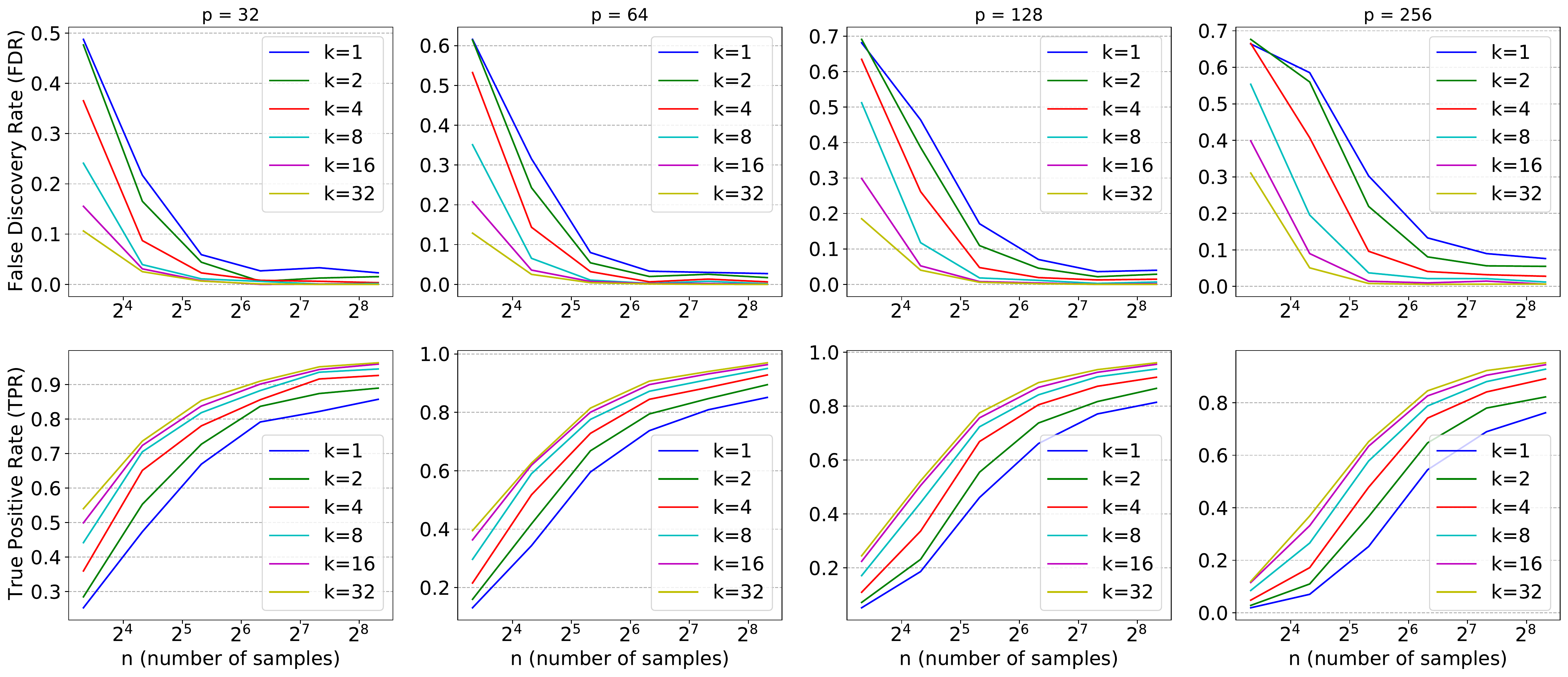}\\
    \vspace{-1mm}
    \caption{False discovery rate (FDR) and true positive rate (TPR) of the edges.}
    % \vspace{-4mm}
    \label{fig:k-acc}
\end{figure}

\textbf{Effectiveness of $K$ in structure recovery.} In this experiment, we aim at verifying the effectiveness of the joint estimation algorithm for recovering the structures. For brevity, we
only report the numbers for false discovery rate (FDR) and true positive rate (TPR) of edges in  Fig~\ref{fig:k-acc}, but figures for additional metrics can be found in Appendix~\ref{app:experiments}.  In Fig~\ref{fig:k-acc}, we can observe consistent improvements when increasing the number of tasks $K$. When per task sample size is small, this improvement reveals to be more obvious.

\textbf{Comparison with other joint estimator.} In this experiment, we compare our method MultiDAG with JointGES~\cite{lee2020bayesian} on models with $p=32$ and $p=64$. Results in Table~\ref{tab:comparison} show that two algorithms have similar performance when $K=1$. However, when $K$ increases, our method returns consistently better structures in terms of SHD.

\begin{table}[htb]
\footnotesize
\centering
\caption{Comparison of MultiDAG (MD) and JointGES (JG) in SHD}
\begin{tabular}{l|rrrr|rrrr}
\multicolumn{1}{c}{} & \multicolumn{4}{c}{$p=32$} & \multicolumn{4}{c}{$p=64$} \\
\toprule
& $n=10$ & $n = 20$ & $n=80$ & $n=320$ & $n=10$ & $n = 20$ & $n=80$ & $n=320$ \\
\midrule
MD(k=1) & $39\pm5$ & $25\pm5$ & $10\pm4$ & $6\pm3$ & $104\pm7$ & $77\pm8$ & $29\pm6$ & $19\pm8$ \\
MD(k=2) & $37\pm5$ & $22\pm5$ & $8\pm3$ & $4\pm3$ & $103\pm7$ & $68\pm8$ & $22\pm6$ & $13\pm6$ \\
MD(k=8) & $29\pm5$ & $13\pm3$ & $4\pm2$ & $2\pm1$ & $85\pm7$ & $42\pm6$ & $13\pm3$ & $6\pm3$\\
MD(k=32) & $23\pm4$ & $11\pm3$ & $3\pm2$ & $1\pm1$ & $66\pm5$ & $35\pm5$ & $10\pm3$ & $3\pm2$\\
\midrule
JG(k=1) & $31\pm5$ & $19\pm5$ & $8\pm4$ & $6\pm5$ & $100\pm11$ & $53\pm11$ & $18\pm11$ & $18\pm9$ \\
JG(k=2) & $32\pm4$ & $19\pm5$ & $9\pm5$ & $7\pm5$ & $99\pm10$ & $51\pm10$ & $20\pm9$ & $21\pm10$ \\
JG(k=8) & $30\pm5$ & $19\pm5$ & $12\pm4$ & $10\pm4$ & $82\pm10$ & $42\pm10$ & $20\pm5$ & $27\pm9$ \\
JG(k=32) & $26\pm4$ & $18\pm4$ & $12\pm3$ & $9\pm3$ & $57\pm9$ & $36\pm6$ & $19\pm5$ & $26\pm6$\\
\bottomrule
\end{tabular}
\label{tab:comparison}
\end{table}

\subsection{Recovery of gene regulatory network}

We investigate how our joint estimator works on more realistic models, by conducting a set of experiments on realistic gene expression data generated by SERGIO~\citep{dibaeinia2020sergio}, which models the additive effect of cooperative transcription factor binding across multiple gene regulators in parallel with protein degradation and noisy expression.

\begin{figure}[h]
    \centering
    \begin{tabular}{@{}cccc@{}}
     (a) True: Ecoli 100   & (b) $K=1,n=1000$ & (c)  $K=10, n=100$ & (d) $K=1, n=100$ \\
    \includegraphics[width=0.2\textwidth]{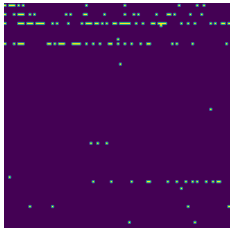}
    & 
    \includegraphics[width=0.2\textwidth]{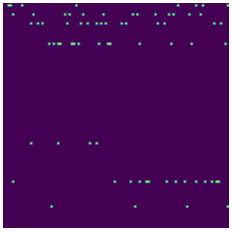}
    &
    \includegraphics[width=0.2\textwidth]{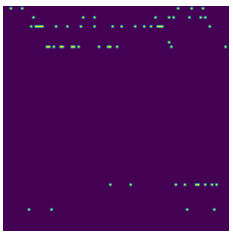}
    &
    \includegraphics[width=0.2\textwidth]{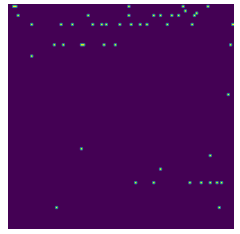}\\
         & FDR: 0.11, TPR: 0.47 & FDR:0.06, TPR:0.41 & FDR: 0.18, TPR: 0.29 \\
    \end{tabular}
    \caption{Visualization of the recovered DAG structures. Each light green colored pixel  at position $(i,j)$ indicates an edge from node $i$ to $j$.  }
    \label{fig:sergio}
\end{figure}

We conduct this experiment on the E. coli 100 network, which includes 100 known genes and 137 known regulatory interactions. To evaluate our algorithm, we generate multiple networks by rearranging and re-weighting 5 edges at random in this network without violating the topological order. We simulate the gene expression from each network using SERGIO with a universal non-cooperative hill coefficient of 0.05, which works well with our linear recovery algorithm.

Fig~\ref{fig:sergio} provides a visual comparison of the recovered structures. It can be seem from the true network that there are a few key transcription factors that highlight several rows in the figure. These transcription factors are better identified by the two structures in (b) and (c), but not that clear in (d). Combining this observation with the more quantitative results in Table~\ref{tab:sergio_table}, we see that the combination $(K=10,n=100)$ achieves comparable performance to $K=1$ with the same total number of samples, and outperforms the single task estimation with $n=100$.

\begin{table}[h]
    \centering
    \begin{tabular}{|c c|c|c|c|c|}
        \hline
        $K$ & $n$ & FDR & TPR & FPR & SHD  \\
        \hline
        1 & 100 & $0.18 \pm 3.8\mathrm{e}{-3}$ & $0.30 \pm 1.2\mathrm{e}{-3}$ & $0.001 \pm 7.3\mathrm{e}{-7}$ & $104.61 \pm 3.2\mathrm{e}{1}$ \\ %& $50.73 \pm 4.5\mathrm{e}{1}$ \\
        10 & 100 & $0.07 \pm 1.2\mathrm{e}{-3}$ & $0.42 \pm 5.1\mathrm{e}{-4}$ & $0.001 \pm 2.8\mathrm{e}{-7}$ & $84.0 \pm 1.5\mathrm{e}{1}$ \\ %& $62.78 \pm 1.8\mathrm{e}{1}$ \\
        1 & 1000 & $0.09 \pm 2.9\mathrm{e}{-3}$ & $0.50 \pm 1.4\mathrm{e}{-3}$ & $0.002 \pm 9.2\mathrm{e}{-7}$ & $76.4 \pm 5.9\mathrm{e}{1}$ \\ % & $75.92 \pm 4.0\mathrm{e}{1}$ \\
        \hline
    \end{tabular}
    \vspace{1mm}
    \caption{
    \small
    Recovery across 25 independent initializations of SERGIO for each experiment. FPR and SHD stand for false positive rate and structural hamming distance, respectively.
    %Training was run with rho=1 and recovered DAGs were thresholded at 0.8
    }
    \label{tab:sergio_table}
\end{table}

\vspace{-3mm}
\section{Conclusion and discussion}
\vspace{-2mm}

In this paper, we have analyzed the behavior of $l_1/l_2$-penalized joint MLE for multiple DAG estimation tasks. Our main result is to show that its performance in recovering the causal order is governed by the sample complexity parameter $\theta(n,K,K',p,s)$ in \eqref{eq:sample-complexity}. %The success recovery depends on whether this parameter is larger than a threshold.
Besides, we have proposed an efficient algorithm for approximating the joint estimator via formulating a novel continuous programming, and demonstrated its effectiveness experimentally.  The current work applies to DAGs that have certain similarity in sparsity pattern. It will be interesting to consider whether the joint estimation without the group-norm (and without the union support assumption) can also lead to similar improvement in causal order recovery. 

\section*{Acknowledgement}

Xinshi Chen is supported by the Google PhD Fellowship. This works is done partially during a visit at MBZUAI. We are grateful for the computing resources provided by the Partnership for an Advanced Computing Environment (PACE) at the Georgia Institute of Technology, Atlanta. We are thankful for PACE Research Scientist Fang (Cherry) Liu's excellent HPC consulting.

\bibliographystyle{unsrt}  
\bibliography{bibfile}

\newpage
\appendix

\section{List of definitions and notations}
For the convenience of the reader, we summarize a list of notations blow.
\begin{enumerate}[leftmargin=1cm]
    \item $\hG_j(\hpi):=[\hvG^{(1)}_j(\hpi),\cdots, \hvG^{(K)}_j(\hpi)]\quad \text{and}\quad
    \tG_j(\hpi):=[\tvG_j^{(1)}(\hpi),\cdots, \tvG_j^{(K)}(\hpi)]$.
    \item For all $k\in[K]$, the eigenvalues of $\Sigma^\k$ are in  $[\Lambda_\tmin, \Lambda_\tmax]$, for some constants $0< \Lambda_\tmin \leq \Lambda_\tmax < \infty$.
    \item $\sigma_\tmax = \sup_{k\in[K],j\in[p],\pi\in\Sp} |\sigma_j^\k(\pi)|$. Note that $\sigma_\tmax^2\leq \Lambda_\tmax$.
    \item $c_\tmax := \sup_{k,j,\pi} |1-\sigma_j^\k(\pi)^2|$. Note that $c_\tmax\leq 1+\sigma_\tmax^2 \leq 1+ \Lambda_\tmax$.
    \item  $\rho=\sup_{k\in[K],j\in[p]}\Sigma^\k_{jj}$. Note that $\rho \leq \Lambda_\tmax$.
    \item $s(\pi):= |S(\pi)|$ where $ S(\pi):=\cup_{k\in[K]} \text{supp}(\tG^\k(\pi))$. $s_0:=s(\pi_0)$. $s:=\sup_{\pi\in\sS_p}s(\pi)$.
    \item $g_\tmax := \sup_{\pi\in\Sp,(i,j)\in S(\pi)} \norm{ \tG_{ij}^\allk (\pi)}_2/\sqrt{K}$.
    \item $g_\tmin := \inf_{(i,j)\in S(\pi_0)}\norm{\tG_{0ij}^\allk}_2/\sqrt{K}$.
    \item $\rsupp_j:=\{i\in[p]: \exists k\in[K]\,s.t.\,\tG^\k_{0ij}\neq 0 \}$.
    \item $r_\tmax:= \sup_{j\in[p]}\big|\rsupp_j\big|$.
    \item $D_\tmax:=\sup_{j\in[p],S=\rsupp_j,k\in[K]}\norm{(\Sigma_{SS}^\k)^{-1}}_\infty$.
    \item $\rho_u: = \sup_{j\in[p],S=\rsupp_j,k\in[K]}\max_{i\in S^c}\rbr{\Sigma_{S^c S^c | S}^\k}_{ii}$.
    \item $S_j(\pi):=\{i: \pi(i)<\pi(j)\}$.
    \item $U_j(\pi):=\cup_{k\in[K]}\text{supp}\big(\tvG_j^\k(\pi)\big)=\{i: \exists k\in[K] \,s.t.\, \tG_{ij}^\k(\pi)\neq 0 \}$.
    \item $\uniond_j:=\sup_{\pi\in\sS_p}|U_j(\pi)|$. $\uniond = \sup_{j\in[p]}\uniond_j$.
\end{enumerate}

\section{Details of Theorem~\ref{thm:order-recovery}: causal order recovery
}
\label{app:thm1}

In Appendix~\ref{app:thm-proof-order-(a)}, we present a general statement of Theorem~\ref{thm:order-recovery} (a) along with its proof. Proof of part (b) in Theorem~\ref{thm:order-recovery} is given in
Appendix~\ref{app:proof-order-fnorm}.

\subsection{Order recovery: proof of Theorem~\ref{thm:order-recovery} (a) (Theorem~\ref{thm:order-recovery-general}) }
\label{app:thm-proof-order-(a)}

Theorem~\ref{thm:order-recovery} (a) states the order recovery guarantee for a specified parameter $\lambda=\sqrt{\frac{p\log p}{n}}$. In the following, we will present a more general statement of Theorem~\ref{thm:order-recovery} (a) that does not specify the choice of $\lambda$, after which we will present the proof.

\begin{theorem}[General statement of Theorem~\ref{thm:order-recovery} (a)]
\label{thm:order-recovery-general}
For any $\delta_1,\delta_2,\delta_3,\delta_4,\delta_5 \in (0,1)$, if the following conditions are satisfied
    \begin{align*}
        K & \leq  \frac{ \delta_2^2\delta_3^2 \delta_4^2\lambda^2  n}{ 64\rho\sigma_\tmax^2},\\
         n & \geq \rbr{\rbr{4(1-\delta_3)\delta_3^{-2}-\delta_5}}^{-1}\rbr{\log K + (\uniond+1)\log p}, \\
        \frac{1}{\eta_w} & > \rbr{\frac{16\sigma_\tmax^8}{\delta_1(1-\delta_1)}+\frac{4\sigma_\tmax^4 c_\tmax}{1-\delta_1}}\frac{2\log p}{nK'} + \frac{4\sigma_\tmax^4g_\tmax}{\delta_1}\frac{\lambda \rbr{s(\pi_0)+\delta_2 s(\hpi)} }{p}\sqrt{\frac{K}{K'{}^2}} \\
         & \quad + \frac{8\sigma_\tmax^6}{\delta_1} \sqrt{\frac{2 \log p}{n}\frac{K-K'}{K'{}^2}},
    \end{align*}
    then $ \hpi = \pi_0$ with probability at least
    \begin{align*}
        1 &- \exp\rbr{-t^*(1-\delta_4)+ (\uniond+2)\log p } - \exp\rbr{-\delta_5 n} -2\exp\rbr{-p\log p},
    \end{align*}
    where $t^*:=  \frac{ \delta_2^2\delta_3^2  \lambda^2  n}{ 16\rho\sigma_\tmax^2}$.
\end{theorem}

% mark: t^* > (d+2) log p

\textbf{Proof outline.}
% Recall that the covariance matrix of each $\rvX^\k$ is
% \begin{align*}
%     &\Sigma^\k = (1-\tG^\k_0)^{-T}(1-\tG^\k_0)^{-1} = (1-\tG^\k(\pi))^{-T}\Omega^\k(\pi)(1-\tG^\k(\pi))^{-1},\\
%     &\Omega^\k(\pi) = \text{diag}\sbr{\sigma_1^{\k }(\pi)^2,\cdots, \sigma_p^{\k}(\pi)^2}.
% \end{align*}
By optimality of the joint estimator,  (for simplicity, we write $\hG^\k:=\hG^\k(\hpi)$)
% \begin{align*}
%     &\sum_{k=1}^K \frac{1}{2n} \|\mX^\k - \mX^\k \hG^\k\|_F^2 + \lambda 
%     \gn{\hG^\allk} \\
%     &\leq \sum_{k=1}^K \frac{1}{2n} \|\mX^\k - \mX^\k \tG^\k_0\|_F^2 + \lambda 
%     \gn{\tG^\allk_0}.
% \end{align*}
% Combining with the following equality
% \begin{align*}
%     \|\mX^\k - \mX^\k \hG^\k\|_F^2 = & \|\mX^\k - \mX^\k \tG^\k(\hpi)\|_F^2 + \|\mX^\k \hG^\k - \mX^\k \tG^\k(\hpi)\|_F^2 \\
%     & - 2\abr{\mX^\k - \mX^\k \tG^\k(\hpi),\mX^\k \hG^\k - \mX^\k \tG^\k(\hpi)}_F,
% \end{align*}
% we can write the optimality condition as
\begin{align*}
    &\sum_{k=1}^K \frac{1}{2n} \|\mX^\k \hG^\k - \mX^\k \tG^\k(\hpi)\|_F^2 + \lambda \gn{\hG^\allk } \\
    \leq &  \underbrace{\sum_{k=1}^K\frac{1}{2n}\rbr{\|\mX^\k - \mX^\k \tG^\k_0\|_F^2-  \|\mX^\k - \mX^\k \tG^\k(\hpi)\|_F^2}}_{(I)} \\
    & +\underbrace{\sum_{k=1}^K\frac{1}{n}\langle \mX^\k - \mX^\k \tG^\k(\hpi),\mX^\k \hG^\k - \mX^\k \tG^\k(\hpi)  \rangle_F}_{(II)} +  \lambda 
    \gn{\tG^\allk_0}.
\end{align*}

The proof is based on a bound for the term (I) and a bound for the term (II). 

To bound (I), we show that the empirical variances of the error terms are close to their expectations, which is achieved mainly by a concentration bound on a linear combination of Chi-squared random variables.

To bound (II), we show the following inequality holds true for all $j\in[p]$ and $\hpi\in\sS_p$ with high probability (where $\te^\k_j(\hpi)$ is the empirical error):
\begin{align*}
     \sup_{\{\beta^\k\in \R^m\}}\frac{1}{n} \sum_{k=1}^K \langle \te^\k_j(\hpi), \mX^\k_{S_j} \beta^\k  \rangle - \frac{\delta}{2n}\sum_{k=1}^K\|\mX^\k_{S_j} \beta^\k \|_2^2-\delta\lambda\|[\beta^{(1)},\cdots,\beta^{(K)}]\|_{l_1/l_2} \leq 0.
\end{align*}
We highlight two technical aspects in bounding (II):
\begin{itemize}[leftmargin=*]
    \item For each fixed $j$ and $\hpi$, the above inequality is proved by showing the \textbf{null-consistency} of $l_1/l_2$-penalized group Lasso problem (see Appendix~\ref{sec:null-consistency}).  Null-consistency means successfully recovering the true linear regression model when the true parameters have null support (all parameters are zeros). Technically, the improvement in sample complexity for recovering multiple DAGs partially comes from the benefit of a larger $K$ for guaranteeing the null-consistency. 
    \item We need to insure the bounds hold uniformly over all permutations $\hpi\in\sS_p$ and $j\in[p]$. To avoid using a naive union bound over $p!$ many permutations, we leverage the sparsity of the graph structures and prove that the number of elements in the set $\{\tG^\allk(\pi): \pi\in\sS_p\}$ can be fewer than $p!$ (see Appendix~\ref{app:invariant}), so that we can take a uniform control over this smaller set instead.
\end{itemize}

We summarize the bounds for (I) and (II) in Lemma~\ref{lm:I} and Lemma~\ref{lm:II}, which can be found in  Appendix~\ref{sec:proof-lm-I} and Appendix~\ref{sec:lm-II}.

\textbf{Detailed proof of Theorem~\ref{thm:order-recovery-general}.} 
Collecting the results in Lemma~\ref{lm:I} and Lemma~\ref{lm:II} and reorganizing the terms in the inequalities, we have the following conclusion.

For any $\delta_1,\delta_2,\delta_3,\delta_4,\delta_5 \in (0,1)$ and $t^*:=  \frac{ \delta_2^2\delta_3^2  \lambda^2  n}{ 16\rho\sigma_\tmax^2}$, if the following conditions are satisfied:
\begin{align*}
    K \leq \frac{\delta_4^2}{4}t^* =  \frac{ \delta_2^2\delta_3^2 \delta_4^2\lambda^2  n}{ 64\rho\sigma_\tmax^2}\\
    \rbr{4(1-\delta_3)\delta_3^{-2}-\delta_5} n \geq  \log K + (\uniond+1)\log p,
\end{align*}
then with probability at least $
    1 - \exp\rbr{-t^*(1-\delta_4)+ (\uniond+2)\log p } - \exp\rbr{-\delta_5 n} -2\exp\rbr{-p\log p}$,
it holds for all $\hpi\in\mathbb{S}_p$ that
\begin{align}
    &  \frac{1-\delta_2}{2n} \sum_{k=1}^K \|\mX^\k \hG^\k - \mX^\k \tG^\k(\hpi)\|_F^2 + \lambda 
    \|\hG^\allk\|_{l_1/l_2}  \nonumber \\
    &\quad + \frac{\delta_1}{4\sigma_\tmax^4}  \sum_{k=1}^{K'}\sum_{j=1}^p\rbr{\sigma_j^\k(\hpi)^2 - \sigma_j^\k(\pi_0)^2 }^2  \nonumber \\
     \leq  & \rbr{\frac{4\sigma_\tmax^4}{1-\delta_1}+ 2 \sigma_\tmax^2}\frac{2p\log p}{n}+ 2\sigma_\tmax^2 \sqrt{\frac{(K-K')2p^2 \log p}{n}} \\
     &\quad + \delta_2 \lambda\|\hG^\allk- \tG^\allk(\hpi)\|_{l_1/l_2} + \lambda\|\tG_0^\allk\|_{l_1/l_2} \nonumber \\
    \leq & \rbr{\frac{4\sigma_\tmax^4}{1-\delta_1}+2 \sigma_\tmax^2}\frac{2p\log p}{n}+ 2\sigma_\tmax^2 \sqrt{\frac{(K-K')2p^2 \log p}{n}} \nonumber \\
    & \quad + \delta_2 \lambda\|\hG^\allk\|_{l_1/l_2} + \delta_2 \lambda\|\tG^\allk(\hpi)\|_{l_1/l_2} + \lambda\|\tG_0^\allk\|_{l_1/l_2}. \label{eq:upper-bound}
\end{align}
Suppose $\hpi\neq \pi_0$. Condition~\ref{cond:iden} implies 
\begin{align*}
    \frac{\delta_1}{4\sigma_\tmax^4}  \frac{pK'}{\eta_w} 
    \leq & \rbr{\frac{4\sigma_\tmax^4}{1-\delta_1}+2 \sigma_\tmax^2}\frac{2p\log p}{n} +\lambda \|\tG_0^\allk\|_{l_1/l_2}  + \delta_2 \lambda\|\tG^\allk(\hpi) \|_{l_1/l_2}\\
    & \quad +  2\sigma_\tmax^2 \sqrt{\frac{(K-K')2p^2 \log p}{n}}.
\end{align*}
Divide both sides by $pK'$, it implies
\begin{align*}
    \frac{\delta_1}{4\sigma_\tmax^4}  \frac{1}{\eta_w} \leq &\rbr{\frac{4\sigma_\tmax^4}{1-\delta_1}+2 \sigma_\tmax^2}\frac{2\log p}{nK'} +\frac{\lambda \|\tG_0^\allk\|_{l_1/l_2}}{pK'}  + \delta_2 \frac{\lambda\|\tG^\allk(\hpi) \|_{l_1/l_2}}{pK'}\\
    & \quad +  2\sigma_\tmax^2 \sqrt{\frac{(K-K')2 \log p}{nK'{}^2}} \\
    \leq & \rbr{\frac{4\sigma_\tmax^4}{1-\delta_1}+c_\tmax}\frac{2\log p}{nK'} + \frac{\lambda \rbr{s(\pi_0)+\delta_2 s(\hpi)} g_\tmax \sqrt{K} }{pK'}\\
    & \quad +  2\sigma_\tmax^2 \sqrt{\frac{(K-K')2 \log p}{nK'{}^2}}.
\end{align*}
The last inequality uses the fact that $\|\tG^\allk(\pi) \|_{l_1/l_2} \leq s(\pi)\sqrt{K} g_\tmax$. It contradicts with the condition
\begin{align*}
    \frac{1}{\eta_w} & > \rbr{\frac{16\sigma_\tmax^8}{\delta_1(1-\delta_1)}+\frac{4\sigma_\tmax^4 c_\tmax}{1-\delta_1}}\frac{2\log p}{nK'} + \frac{4\sigma_\tmax^4g_\tmax}{\delta_1}\frac{\lambda \rbr{s(\pi_0)+\delta_2 s(\hpi)} }{p}\sqrt{\frac{K}{K'{}^2}} \\
    & \quad + \frac{8\sigma_\tmax^6}{\delta_1} \sqrt{\frac{2 \log p}{n}\frac{K-K'}{K'{}^2}}.
\end{align*} Therefore, $\hpi\in\Pi_0$.

% Take $\lambda = \sqrt{\frac{p\log p}{n}}$. Then $t^* =\frac{ \delta_2^2\delta_3^2 p\log p }{ 16\rho\sigma_\tmax^2}$. Assume 
% \begin{align*}
%     &K \leq \frac{\delta_4^2}{4}t^* =  \frac{ \delta_2^2\delta_3^2 \delta_4^2 p\log p }{ 64\rho\sigma_\tmax^2}.
% \end{align*}
% Then it holds that
% \begin{align*}
%     \frac{1-\delta_1}{4\sigma_\tmax^4}  \frac{1}{\eta_w} \leq & \rbr{\frac{4\sigma_\tmax^4}{1-\delta_1}+c_\tmax}\frac{2\log p}{nK} + g_\tmax\frac{s(\pi_0)+\delta_2 s(\hpi)}{\sqrt{p}}\sqrt{\frac{ \log p }{nK}},
% \end{align*}
% which implies
% \begin{align*}
%       \frac{1}{\eta_w} \leq & \rbr{1+\frac{4c_\tmax \sigma_\tmax^4}{1-\delta_1} }\frac{2\log p}{nK} + g_\tmax\frac{s(\pi_0)+\delta_2 s(\hpi)}{\sqrt{p}}\sqrt{\frac{ \log p }{nK}}.
% \end{align*}

Theorem~\ref{thm:order-recovery} (a) is straightforward by taking $\lambda = \sqrt{p\log p / n}$.

\subsection{Key lemmas for proving Theorem~\ref{thm:order-recovery} (a)}

\subsubsection{Lemma~\ref{lm:I}: Analysis of (I)}
\label{sec:proof-lm-I}

\begin{lemma}
\label{lm:I}
Denote $\sigma_\tmax = \sup_{k\in[K],j\in[p],\pi\in\Sp} |\sigma_j^\k(\pi)|$.
With probability at least $1-2e^{-p\log p }$, it holds for any $\delta_1\in(0,1)$ and any permutations $\hpi \in \mathbb{S}_p$ that, 
\begin{align*}
    &\sum_{k=1}^K\frac{1}{2n}\rbr{\|\mX^\k - \mX^\k \tG^\k_0\|_F^2-  \|\mX^\k - \mX^\k \tG^\k(\hpi)\|_F^2} 
    \\
    \leq & -\frac{\delta_1}{4\sigma_\tmax^4} \sum_{k=1}^{K'}\sum_{j=1}^p \rbr{ \sigma_j^\k(\pi_0)^2 - \sigma_j^\k(\hpi)^2}^2 \\
    & \quad + \rbr{\frac{\sigma_\tmax^4}{1-\delta_1}+2\sigma_\tmax^2}\frac{2p\log p}{n} + 2\sigma_\tmax^2 \sqrt{\frac{(K-K')2p^2 \log p}{n}}.
\end{align*}
\end{lemma}

We now state the proof of this Lemma. 
Denote the $j$-th column of $\tG^\k (\pi)$ as $\tvG^\k_j(\pi)$, and the noise as
\begin{align}
    \te^\k_j(\pi):= \mX_j^\k - \mX^\k  \tvG^\k_j(\pi) \in \R^n. \label{eq:sample-error}
\end{align}
Then we can rewrite the term (I) as follows.
\begin{align*}
    & (I) = \sum_{k=1}^K\frac{1}{2n}\rbr{\|\mX^\k - \mX^\k \tG^\k_0\|_F^2-  \|\mX^\k - \mX^\k \tG^\k(\hpi)\|_F^2} \\
    & = \frac{1}{2}\sum_{k=1}^K \sbr{ \sum_{j=1}^p \frac{\frac{1}{n}\|\te_j^\k(\hpi)\|_2^2}{\sigma_j^\k(\hpi)^2} \sigma_j^\k(\pi_0)^2 - \sum_{j=1}^p \frac{1}{n}\|\te_j^\k(\hpi)\|_2^2} \\
    & = \frac{1}{2}\sum_{k=1}^K  \sum_{j=1}^p \rbr{\sigma_j^\k(\pi_0)^2-\sigma_j^\k(\hpi)^2 }\rbr{\frac{\frac{1}{n}\|\te_j^\k(\hpi)\|_2^2}{\sigma_j^\k(\hpi)^2} - 1} +  \frac{1}{2}\sum_{k=1}^K  \sum_{j=1}^p \rbr{\sigma_j^\k(\pi_0)^2-\sigma_j^\k(\hpi)^2} \\
    & \leq \frac{1}{2}\sum_{k=1}^{K}  \sum_{j=1}^p \rbr{\sigma_j^\k(\pi_0)^2-\sigma_j^\k(\hpi)^2 }\rbr{\frac{\frac{1}{n}\|\te_j^\k(\hpi)\|_2^2}{\sigma_j^\k(\hpi)^2} - 1}  \\
    & \quad -\frac{1}{4\sigma_\tmax^4} \sum_{k=1}^{K'}  \sum_{j=1}^p\rbr{\sigma_j^\k(\hpi)^2 - \sigma_j^\k(\pi_0)^2}^2
\end{align*}
The last inequality holds because for $k=1,\cdots,K'$,
$
    \sum_{j=1}^p \rbr{\sigma_j^\k(\pi_0)^2-\sigma_j^\k(\hpi)^2} \leq -\frac{1}{2\sigma_\tmax^4} \sum_{j=1}^p\rbr{\sigma_j^\k(\hpi)^2 - \sigma_j^\k(\pi_0)^2}^2$, 
and that for $k>K'$, $
    \sum_{j=1}^p \rbr{\sigma_j^\k(\pi_0)^2-\sigma_j^\k(\hpi)^2} \leq 0$. 
Then we bound the first term using the concentration bound on Chi-squared random variables.
\begin{align*}
     & \sum_{k=1}^K  \sum_{j=1}^p \rbr{\sigma_j^\k(\pi_0)^2-\sigma_j^\k(\hpi)^2 }\rbr{\frac{\frac{1}{n}\|\te_j^\k(\hpi)\|_2^2}{\sigma_j^\k(\hpi)^2} - 1} \\
     & \deq \sum_{k=1}^K  \sum_{j=1}^p \rbr{\sigma_j^\k(\pi_0)^2-\sigma_j^\k(\hpi)^2 }\rbr{\frac{1}{n} \xi_j^2 -1}  = \frac{1}{n}\sum_{k=1}^K  \sum_{j=1}^p \rbr{\sigma_j^\k(\pi_0)^2-\sigma_j^\k(\hpi)^2 }\rbr{\xi_j^2 -n}, \\
     &\text{where }\xi_j^2\sim \chi^2(n) \text{ are i.i.d. Chi-squared random variables of degree $n$.}
\end{align*}
By Lemma~\ref{lm:chi-square}, for any fixed $\hpi \in \Sp$ and for any $t>0$, it holds with probability at least $1-e^{-t}$ that 
\begin{align*}
    & \frac{1}{n}\sum_{k=1}^{K'}  \sum_{j=1}^p \rbr{\sigma_j^\k(\pi_0)^2-\sigma_j^\k(\hpi)^2 }\rbr{\xi_j^2 -n} \\
    &\leq 2 \sqrt{\frac{\sum_{k=1}^{K'}\sum_{j=1}^p \rbr{\sigma_j^\k(\pi_0)^2 - \sigma_j^\k(\hpi)^2}^2}{n}}\sqrt{t}  + \frac{2 \sigma_\tmax^2}{n}t\\
    &\leq \delta \sum_{k=1}^{K'}\sum_{j=1}^p \rbr{\sigma_j^\k(\pi_0)^2 - \sigma_j^\k(\hpi)^2}^2 + \rbr{\frac{1}{\delta}+2\sigma_\tmax^2}\frac{t}{n},
\end{align*}
The last inequality holds because $2ab \leq \delta a^2 + \frac{1}{\delta} b^2$ for any $\delta >0$. Now it remains to take a union bound over the permutation $\hpi\in\sS_p$. There are $p!$ many permutations. Take an uniform control over all possible $\hpi\in\sS_p$. It implies that with probability at least $1-{(p!)}e^{-t}$, the above inequality holds for all $\hpi\in\sS_p$. Equivalently, we can say it holds with probability at least $1-e^{-t}$ that it holds for all $\hat{\pi}$ that
\begin{align*}
    &\frac{1}{n}\sum_{k=1}^{K'}  \sum_{j=1}^p \rbr{\sigma_j^\k(\pi_0)^2-\sigma_j^\k(\hpi)^2 }\rbr{\xi_j^2 -n} \\
    &\leq \delta \sum_{k=1}^{K'}\sum_{j=1}^p \rbr{\sigma_j^\k(\pi_0)^2- \sigma_j^\k(\hpi)^2}^2 + \rbr{\frac{1}{\delta}+2\sigma_\tmax^2}\frac{t+p\log p}{n}.
\end{align*}
For the non-identifiable models, we can use Lemma~\ref{lm:chi-square} in a similar way to obtain that with probability at least $1-e^{-t}$, the following holds for all $\hat{\pi}$,
\begin{align*}
    & \frac{1}{n}\sum_{k={K'+1}}^{K}  \sum_{j=1}^p \rbr{\sigma_j^\k(\pi_0)^2-\sigma_j^\k(\hpi)^2 }\rbr{\xi_j^2 -n} \\
    & \leq 2 \sqrt{\frac{\sum_{k={K'+1}}^{K}\sum_{j=1}^p \rbr{\sigma_j^\k(\pi_0)^2 - \sigma_j^\k(\hpi)^2}^2}{n}}\sqrt{t}  + \frac{2 \sigma_\tmax^2}{n}t\\
    & \leq 2\sigma_\tmax^2 \sqrt{\frac{(K-K')p(t+p\log p)}{n}}   + \frac{2 \sigma_\tmax^2}{n}(t+p\log p).
\end{align*}
Putting the above results back into the term (I), taking $\delta' = 2 \delta$, and taking $t=p\log p$, we have  with probability at least $1-2e^{-p\log p}$ that
\begin{align*}
    (I) & \leq \frac{\delta'}{4} \sum_{k=1}^{K'}\sum_{j=1}^p \rbr{\sigma_j^\k(\pi_0)^2 - \sigma_j^\k(\hpi)^2}^2 -\frac{1}{4\sigma_\tmax^4} \sum_{k=1}^{K'}  \sum_{j=1}^p\rbr{\sigma_j^\k(\hpi)^2 - \sigma_j^\k(\pi_0)^2}^2 \\
    & \quad + \rbr{\frac{1}{\delta'}+2\sigma_\tmax^2}\frac{2p\log p}{n}  + 2\sigma_\tmax^2 \sqrt{\frac{(K-K')2p^2 \log p}{n}}
\end{align*}
Finally, take $\delta_1 = 1- \sigma_\tmax^4 \delta' $ so that $\delta' = \frac{1}{ \sigma_\tmax^4} (1-\delta_1)$. Then for any $\delta_1\in(0,1)$, the following inequality holds with probability at least $1-e^{-p\log p}$ for all $\hpi\in\sS_p$:
\begin{align*}
    (I) \leq & -\frac{\delta_1}{4\sigma_\tmax^4} \sum_{k=1}^{K'}\sum_{j=1}^p \rbr{ \sigma_j^\k(\pi_0)^2 - \sigma_j^\k(\hpi)^2}^2 \\
    & \quad + \rbr{\frac{\sigma_\tmax^4}{1-\delta_1}+2\sigma_\tmax^2}\frac{2p\log p}{n} + 2\sigma_\tmax^2 \sqrt{\frac{(K-K')2p^2 \log p}{n}}.
\end{align*}

\subsubsection{Lemma~\ref{lm:II}: Analysis of (II)}
\label{sec:lm-II}

\begin{lemma}\label{lm:II}
Denote $\rho :=\sup_{k\in[K],j\in[p]}\Sigma^\k_{jj}$ and $\sigma_\tmax:= \sup_{k\in[K],j\in[p],\pi\in\Sp} |\sigma_j^\k(\pi)|$. 
For any $\delta_2,\delta_3\in(0,1)$,
assume $\lambda$ satisfies $t^*:=\frac{\delta_2^2\delta_3^2\lambda^2 n}{16\rho\sigma_\tmax^2}>K$. With probability at least
\begin{align*}
    1 &- \exp\rbr{-t^*\sbr{1-2\sqrt{\frac{K}{t^*}}}+ (\uniond+2)\log p } - \exp\rbr{-\frac{4(1-\delta_3)}{\delta_3^2}n + \log K + (\uniond+1)\log p},
\end{align*}
the following inequality holds true:
\begin{align*}
     &\frac{1}{2n}\sum_{k=1}^K\langle \mX^\k - \mX^\k \tG^\k(\hpi),\mX^\k \hG^\k - \mX^\k \tG^\k(\hpi) \rangle \\
     &\leq  \frac{\delta_2}{2n}\sum_{k=1}^K\|\mX^\k \hG^\k - \mX^\k \tG^\k(\hpi) \|_F^2  + \delta_2 \lambda\|\hG^\allk- \tG^\allk(\hpi)\|_{l_1/l_2}.
\end{align*}
\end{lemma}

We now state the proof of this Lemma. 
To show the inequality in Lemma~\ref{lm:II} holds true, it is sufficient to show the following inequality holds true for all $j$ and $\hpi$:
\begin{align}
    \frac{1}{n} \sum_{k=1}^K \langle \te^\k_j(\hpi), \mX^\k\rbr{ \hvG^\k_j(\hpi)-\tvG^\k_j(\hpi)}  \rangle_F \leq & \frac{\delta}{2n}\sum_{k=1}^K\|\mX^\k\rbr{ \hvG^\k_j(\hpi)-\tvG^\k_j(\hpi)} \|_2^2 \nonumber \\
         & -\delta \lambda\|\hvG^\allk_j(\hpi)-\tvG^\allk_j(\hpi)\|_{l_1/l_2}, \label{eq:1}
\end{align}
where we denote
\begin{align*}
    \hG_j(\hpi):=[\hvG^{(1)}_j(\hpi),\cdots, \hvG^{(K)}_j(\hpi)]\quad \text{and}\quad
    \tG_j(\hpi):=[\tvG_j^{(1)}(\hpi),\cdots, \tvG_j^{(K)}(\hpi)].
\end{align*} 
\underline{Now consider a fixed $j$ and a fixed $\hpi$}. Recall $S_j(\hpi)$ which denotes the set of ancestors of the node $j$ specified by the permutation $\hpi$, and let $m=|S_j(\hpi)|\in[0,p-1]$ be its cardinality. Let $\mX^\k_{S_j}=\mX^\k|_{S_j(\hpi)}$ denote the submatrix of $\mX^\k$ whose column indices are in $S_j(\hpi)$.
We define the event
\begin{align*}
    &\gE(\delta,\lambda;\te^\k_j(\hpi)):=\\
    &\cbr{ \sup_{\{\beta^\k\in \R^m\}}\frac{1}{n} \sum_{k=1}^K \langle \te^\k_j(\hpi), \mX^\k_{S_j} \beta^\k  \rangle - \frac{\delta}{2n}\sum_{k=1}^K\|\mX^\k_{S_j} \beta^\k \|_2^2-\delta\lambda\|[\beta^{(1)},\cdots,\beta^{(K)}]\|_{l_1/l_2} \leq 0}.
\end{align*}
It's easy to see that with probability at least $\Pr\sbr{\gE(\delta,\lambda;\te^\allk_j(\hpi))}$, the inequality in \eqref{eq:1} holds true. Therefore, we need to derive the probability of the joint event $\cap_{j\in[p],\hpi\in\Sp}\gE(\delta,\lambda;\te^\allk_j(\hpi))$ in this proof.
Observe that:
\begin{align*}
   &\gE(\delta,\lambda;\te^\allk_j(\hpi)) \subseteq \\
   &\cbr{ \sup_{\{\beta^\k\in \R^m\}} \frac{1}{2n}\sum_{k=1}^K\|\frac{\te^\k_j(\hpi)}{\delta} \|_2^2 -\frac{1}{2n} \sum_{k=1}^K \| \frac{\te^\k_j(\hpi)}{\delta} - \mX^\k_{S_j} \beta^\k\|_2^2 -\lambda\|[\beta^{(1)},\cdots,\beta^{(K)}]\|_{l_1/l_2} \leq 0}\\
   &= \cbr{\vzero \in \argmin_{\{\beta^\k\in \R^m\}}\frac{1}{2n} \sum_{k=1}^K \| \frac{\te^\k_j(\hpi)}{\delta} - \mX^\k_{S_j} \beta^\k\|_2^2 + \lambda\|[\beta^{(1)},\cdots,\beta^{(K)}]\|_{l_1/l_2} }
\end{align*}
Therefore, we resort to bound the probability of the above event, which is the \textbf{null-consistency} of $l_1/l_2$-penalized group Lasso problem. We present the null-consistency analysis by  Lemma~\ref{lm:null-consistency}, and its proof is given in Sec~\ref{sec:null-consistency}.

Note that in the event $\gE(\delta,\lambda;\te^\allk_j(\hpi))$, the variance is
$
  \frac{\te^\k_j(\hpi)}{\delta} \deq \vw_k \quad $ with
\begin{align*}
    \vw_k\sim \gN\rbr{0, \rbr{\frac{\sigma_j^\k(\hpi)}{\delta}}^2 I_n}.
\end{align*}
Therefore, we can take $\sigma_0$ in Lemma~\ref{lm:null-consistency} to be $\sigma_0 = \frac{\sigma_\tmax}{\delta}$, which implies  for any $\delta'\in(0,1)$, if 
\begin{align*}
    t^*:=\frac{\delta'{}^2\delta^2\lambda^2 n}{16\rho\sigma_\tmax^2}>K,
\end{align*}
then
\begin{align*}
    \Pr\sbr{\gE(\delta,\lambda;\te^\allk_j(\hpi))} \geq  1- p\exp\rbr{-t^*\sbr{1-2\sqrt{\frac{K}{t^*}}}} - K \exp\rbr{-\frac{4(1-\delta')}{\delta'{}^2}n}.
\end{align*}
Now what remains is to take a uniform control over all events $\cap_{j\in[p],\hpi\in\Sp}\gE(\delta,\lambda;\te^\allk_j(\hpi))$. A naive way is to enumerate over all permutations $\hpi$ and all $j$ which will constitute $p\cdot{p!}$ events. However, recall $\te^\k_j(\pi):= \mX_j^\k - \mX^\k  \tvG^\k_j(\pi)$. Then it is enough to take a uniform control over the set $\{\tvG_j^\allk(\pi): \pi\in\sS_p, j\in[p]\}$. By~\eqref{eq:invariant-size}, this set contains at most $p\cdot p^{\uniond}$ many elements. Therefore,
\begin{align*}
    &\Pr\sbr{\cap_{j\in[p],\hpi\in\Sp}\gE(\delta,\lambda;\te^\allk_j(\hpi))} \\
    &\geq  1- p^2 p^\uniond \exp\rbr{-t^*\sbr{1-2\sqrt{\frac{K}{t^*}}}}  - Kp \cdot p^\uniond \exp\rbr{-\frac{4(1-\delta')}{\delta'{}^2}n}\\
    & \geq 1- \exp\rbr{-t^*\sbr{1-2\sqrt{\frac{K}{t^*}}}+ (\uniond+2)\log p} - \exp\rbr{-\frac{4(1-\delta')}{\delta'{}^2}n + \log K + (\uniond+1)\log p}
\end{align*}
which implies \eqref{eq:1} holds with the above probability.

\subsubsection{Lemma~\ref{lm:null-consistency}: Null Consistency}
\label{sec:null-consistency}

\begin{lemma}[Null-consistency]
\label{lm:null-consistency} Let $S\subseteq [p]$ be a set of $m$ indices.  
Consider the following linear regression model with zero vector as the true parameters:
\begin{align*}
    \vY^\k = \Xk_S \vzero + \vW^\k,\quad \text{for }k\in[K]
\end{align*}
where $\vY^\k= \vW^\k\in\R^n$, $\Xk_S\in\R^{n\times m}$ and  $\vzero \in\R^m$. Assume that for each $k$, the row vectors of $\Xk$ are i.i.d. sampled from $\gN(0, \Sigma^\k)$ and the noise is sampled from $\vw^\k \sim \gN(0, \sigma_W^{\k 2}I_n)$. Denote 
$\rho := \max_{k\in[K]}\Sigma_{jj}^\k \quad \text{and}\quad \sigma_0 := \max_{k\in[K]} \sigma_W^\k$. Consider the following \textbf{$l_1/l_2$-regularized Lasso problem}:
\begin{align}
    \hB = \argmin_{B\in\R^{m\times K}} \frac{1}{2n}\sum_{k=1}^K \|\vY^\k-\Xk_S\vB^\k\|_2^2 + \lambda \|B \|_{l_1/l_2},
    \label{eq:joint-estimator2}
\end{align}
where $B = [\vB^{(1)}, \cdots, \vB^\k]$. For any $\delta\in(0,1)$, if 
\begin{align*}
    t^* =  \frac{\delta^2\lambda^2 n}{16\rho\sigma_0^2}>K,
\end{align*}
then with probability at least
\begin{align*}
    1- m\exp\rbr{-t^*\sbr{1-2\sqrt{\frac{K}{t^*}}}} - K \exp\rbr{-\frac{4(1-\delta)}{\delta^2}n},
\end{align*}
$\hB = \vzero$ is an optimal solution to the problem in \eqref{eq:joint-estimator2}. 
\end{lemma}

The proof of this lemma is stated below, in which we simply use the notation $\mX^\k$ to replace $\mX^\k_S$.
\begin{lemma}
Suppose there exists a primal-dual pair $(\hB, \hZ) \in \R^{m\times K} \times \R^{m\times K}$ which satisfies the following conditions:
\begin{subequations}
\label{eq:conditions}
\begin{align}
     \hZ & \in \partial \|\hB\|_{l_1/l_2}, \\
     -\frac{1}{n} \mX^{\k \top}\rbr{\vy^\k - \mX^\k \hvB^\k} + \lambda \hvZ^\k & = 0, \quad \forall k \in [K], \label{eq:gradient-zero}\\
     \|\hZ\|_{l_\infty/l_2} & < 1,
\end{align}
\end{subequations}
where $[\hvB^{(1)},\cdots,\hvB^{(K)}]$ are the columns of $\hB$ and $[\hvZ^{(1)},\cdots, \hvZ^{(K)}]$ are the columns of $\hZ$. Then $\vzero$ is the solution to the problem in \eqref{eq:joint-estimator} and it is the only solution.
\end{lemma}
\begin{proof}
Straightforward by Lemma 1 in \citep{wang2014collective}.
\end{proof}
Therefore, to show $\vzero \in \argmin_{B\in\R^{m\times K}} \frac{1}{2n}\sum_{k=1}^K \|\vY^\k-\Xk\vB^\k\|_2^2 + \lambda \|B \|_{l_1/l_2}$, it is sufficient to show the existence of $(\hB, \hZ)$ which satisfies the conditions in \eqref{eq:conditions}. We construct such a pair by the following definitions:
\begin{align}
    \hB  & := \vzero, \\
    \hvZ^\k & := \frac{1}{\lambda n}\mX^{\k \top}\vy^\k. \label{eq:def-hZ}
\end{align}
Clearly, they satisfy \eqref{eq:gradient-zero}. If we can show $\|\hZ \|_{l_\infty/l_2} < 1$, then \eqref{eq:conditions} holds. Therefore, in this proof, the main goal is the analyze  $\Pr\sbr{|\hZ \|_{l_\infty/l_2} <1 }$.

Denote the row vectors of $\hZ$ as $\hZ_j := [\hvZ_j^{(1)},\cdots, \hvZ_j^{(K)}]$. Then
$ \|\hZ\|_{l_\infty/l_2} = \max_{j\in [m]} \|\hZ_j\|_2$. By definition in \eqref{eq:def-hZ},
\begin{align*}
    \hvZ_{j}^\k & =\frac{1}{\lambda n}\Xk_j{}^\top \vy^\k = \frac{1}{\lambda n}\Xk_j{}^\top \vw^\k.
\end{align*}
Since the linear combination of Gaussian distribution is still Gaussian, then given $\vw^\allk$, the variable $ \|\hZ_j\|_2^2$ is equivalent to a   Chi-squared random variable in distribution:
\begin{align*}
   \|\hZ_j\|_2^2 \mid \vw^\allk & = \frac{1}{\lambda^2 n^2 } \sum_{k=1}^K \rbr{\Xk_j{}^\top \vw^\k}^2 \mid \vw^\allk  \\
   & \deq \frac{1}{\lambda^2 n^2 }   \sum_{k=1}^K  \Sigma_{jj}^\k\|\vW^\k\|_2^2 \xi_{jk}^2 \quad \text{where}~\xi_{jk}\sim \gN(0,1) \\
    &\leq \frac{1}{\lambda^2 n^2 }  \max_{k\in[K]}\Sigma_{jj}^\k  \max_{k\in[K]}\|\vW^\k\|_2^2\sum_{k=1}^K \xi_{jk}^2 
\end{align*}
By Lemma~\ref{lm:con-chi}, for any $\delta>0$, 
\begin{align*}
    \Pr\sbr{\max_{k\in[K]}\|\vw^\k\|_2^2\geq \sigma_W^{\k 2} 2n(1+\delta)} \leq K \exp\rbr{-n(1+\delta)\sbr{ 1-2\sqrt{\frac{1}{1+\delta}}}}.
\end{align*}
Therefore, for all $\delta>0$,
\begin{align*}
    \Pr\sbr{\max_j\|\hZ_j\|_2 < 1} \geq \Pr\sbr{\max_{j\in[p]} \sum_{k=1}^K \xi_{jk}^2 < \frac{\lambda^2 n}{ 2(1+\delta)\rho  \sigma_0^2 }} \Pr\sbr{\max_{k}\|\vw^\k\|_2^2 < 2n(1+\delta)}
\end{align*}
where
\begin{align*}
    \rho := \max_{k\in[K]}\Sigma_{jj}^\k \quad \text{and}\quad \sigma_0 := \max_{k\in[K]} \sigma_W^\k.
\end{align*}
Take $t^* = \frac{\lambda^2 n}{ 4(1+\delta)\rho\sigma_0^2} $, then if $t^* > K $, we have 
\begin{align*}
    \Pr\sbr{\max_{j\in[p]} \sum_{k=1}^K \xi_{jk}^2 < 2t^*} \geq 1- p\exp\rbr{-t^*\sbr{1-2\sqrt{\frac{K}{t^*}}}}.
\end{align*}
Rewrite $\delta=\frac{4}{\delta'{}^2} -1$ for some $\delta'\in (0,1)$ so that $1+\delta=\frac{4}{\delta'{}^2}>4$. If $\lambda$ is taken to be some value that satisfies the condition
\begin{align*}
    t^* =  \frac{\delta'{}^2\lambda^2 n}{16\rho\sigma_0^2}>K,
\end{align*}
then
\begin{align*}
    \Pr\sbr{\max_j\|\hZ_j\|_2 < 1} \geq 1- p\exp\rbr{-t^*\sbr{1-2\sqrt{\frac{K}{t^*}}}} - K \exp\rbr{-\frac{4(1-\delta')}{\delta'{}^2}n}.
\end{align*}

\subsection{Proof of error in F-norm}
\label{app:proof-order-fnorm}
Denote the error vector as $
    \Delta^\k_j:=\hvG^\k_j(\hpi)-\tvG^\k_j(\hpi)$. 
Then
\begin{align*}
    \sum_{k=1}^K \frac{1}{2n} \|\mX^\k \hG^\k - \mX^\k \tG^\k(\hpi)\|_F^2 = \sum_{k=1}^K \sum_{j=1}^p\frac{1}{2n} \|\mX^\k \Delta^k_j\|_2^2
\end{align*}

By Theorem 7.3 in \citep{van2013ell_}, with probability at least $1-\exp(-\log p- \log K)$, it holds for all $k\in[K]$ and $j\in[p]$ that
\begin{align*}
    \frac{1}{\sqrt{n}} \|\mX^\k \Delta^\k_j \|_2 &\geq \rbr{ 3/4 \Lambda_\tmin  - 3\sigma_\tmax \sqrt{\frac{\hat{d}\rbr{\log p + \log K}}{n}} - \sqrt{\frac{4(\log p + \log K)}{n}}} \|\Delta^\k_j\|_2\\
    &\geq \rbr{ 3/4 \Lambda_\tmin  - 3\sigma_\tmax \sqrt{\frac{\hat{d} \rbr{\log p + \log K}}{n}} -c } \|\Delta^\k_j\|_2
\end{align*}
where $\hat{d}: =\sup_{j,k}\|\Delta^\k_j\|_0$. %\sup_{k\in[K],j\in[p]} 
If the sample size $n$ satisfies the condition with a suitable constant $\kappa(\Lambda_\tmin)$:
\begin{align*}
    n \geq \kappa(\Lambda_\tmin)\hat{d}\rbr{\log p+ \log K},
\end{align*}
then $ \frac{1}{\sqrt{n}} \|\mX^\k \Delta^\k_j \|_2 \geq \kappa'(\Lambda_\tmin) \|\Delta^\k_j\|_2$ for some constant $\kappa'(\Lambda_\tmin) $. Therefore, 
\begin{align*}
    &\frac{1}{K}\sum_{k=1}^K\sum_{j=1}^p \|\Delta^\k_j\|_2^2 \leq \frac{2}{\kappa'(\Lambda_\tmin)^2} \sum_{k=1}^K \frac{1}{2nK} \|\mX^\k \hG^\k - \mX^\k \tG^\k(\hpi)\|_F^2 \\
    & \leq \frac{2}{\kappa'(\Lambda_\tmin)^2} \rbr{\kappa(\sigma_\tmax) \frac{p\log p}{nK } + c g_\tmax \frac{s_0 \lambda}{\sqrt{K}}},
\end{align*}
which implies
\begin{align*}
    &\frac{1}{K}\sum_{k=1}^K\sum_{j=1}^p \|\Delta^\k_j\|_2^2  = \gO\rbr{ \frac{p\log p}{nK}+ \frac{s_0\lambda}{\sqrt{K}} } = \gO\rbr{ \frac{s_0\lambda}{\sqrt{K}}}.
\end{align*}
The last equation holds for the case when $\lambda = \sqrt{\frac{p\log p}{nK}}$.

\section{Proof of Theorem~\ref{thm:support}: Support Recovery}
\label{app:thm2}

We are interested in showing that the support union of $\hG^\allk$ is the same as that of $\tG^\allk_0$. To prove this, we can equivalently show the support union of 
$\hvG_j^\allk$ is the same as that of $\tvG_j^\allk$ for any $j\in[p]$. 

Now we state the proof of Theorem~\ref{thm:support}.

\begin{proof}

Given a permutation $\pi_0\in\Pi_0$, the DAG structure learning problem is equivalent to solving $p$ separate group Lasso problems, where for each $j$, the following $l_1/l_2$-penalized group Lasso is solved:
\begin{align}
    \hvG_j^\allk|_{S_j(\pi_0)} = \argmin_{B\in \R^{|S_j(\pi_0)|\times K}} \sum_{k=1}^K \frac{1}{2n}\|\mX_j^\k - \rbr{\mX^\k|_{S_j(\pi_0)}} \vB^\k \|_2^2 + \lambda \|B\|_{l_1/l_2},
    \label{eq:mdmr}
\end{align}
where $S_j(\pi_0):=\cbr{i: \pi_0(i)<\pi_0(j)}$, and we denote the columns of $B$ as $B = \sbr{\vB^{(1)},\cdots, \vB^{(K)} }$.  The proof in this section is based on a uniform control over all $j$. For each $j$,
the estimation in form of \eqref{eq:mdmr} is called a \textit{multi-design multi-response} (or multivariate) regression problem, which has been studied in the last decade \citep{obozinski2011support,wang2014collective}. Our support recovery analysis is based on techniques for analyzing multivariate regression problem in existing literature, but careful adaptation is needed to simultaneously handle a set of $p$ problems where $p$ is the dimension of the problem.

More precisely, we present the analysis of multivariate regression problems in Appendix~\ref{app:multivariate-regression}, where the main results are summarized in Theorem~\ref{thm:glasso-linear-support}. Then the results in Theorem~\ref{thm:support} can be obtained with direct computations by applying Theorem~\ref{thm:glasso-linear-support} to $p$ separate problems defined by \eqref{eq:mdmr} and taking a union bound over all $j\in[p]$.

% To summarize, if the following conditions hold
% \begin{align*}
%     n \geq \kappa_6 r_\tmax \log p,\\
%     K \leq c_7 \log p,\\
%     \sqrt{ \frac{8\sigma_\tmax^2  \log p}{\Lambda_\tmin n} } + \frac{2 }{\Lambda_\tmin}\sqrt{\frac{p\log p}{n}\frac{r_\tmax}{K}} = o\rbr{g_\tmin},
% \end{align*}
% then the set of $p$ problems in \eqref{eq:mdmr} lead to a unique solution $\hB$, and that with probability at least
% \begin{align*}
%     %&1 - c_1 pK\exp\rbr{-c_2(n-r_\tmax)}- p\exp\rbr{-c_3 \log p}-  ps\exp\rbr{-c_4K\log p}\\
%     %&\geq 
%     1-  r_\tmax \exp\rbr{-c_5 K\log p}- \exp\rbr{-c_6 \log p} - c_8K\exp\rbr{-c_9(n-r_\tmax-\log p)},
% \end{align*}
% the support union of $\hG^\allk$ is the same as $\tG_0^\allk$, and that
% $\|\hG^\allk - \tG_0^\allk\|_{l_\infty/l_2}/\sqrt{K} = o(g_\tmin)$.
\end{proof}

\section{Support Union Recovery for Multi-Design Multi-response Regression}
\label{app:multivariate-regression}

The analysis in this section can be independent of the other content in this paper. We first introduce the multivariate regression setting and notations below.

\textbf{Problem setting and assumptions.} Consider the following $K$ linear regression models
\begin{align*}
    \vY^\k = \Xk \tvB{}^\k + \vW^\k,\quad \text{for }k=1,\cdots, K,
\end{align*}
where $\vY^\k \in\R^n$, $\Xk\in\R^{n\times p}$, $\tvB{}^\k \in\R^p$, and $\vW^\k\in\R^n$. Assume that for each $k$, the row vectors of $\Xk$ are i.i.d. sampled from $\gN(0, \Sigma^\k)$ and the noise is sampled from $\vw^\k \sim \gN(0, \sigma^{\k 2}I_n)$. Denote $S$ as the support union of true parameters $\{\tvB{}^\k\}_{k\in[K]}$, i.e., $S:=\{j: \exists k\in[K] \,s.t.,\,\tvB_j{}^\k\neq 0\}$, and $s=|S|$ as its size. Note that the $s$ in this section has a different meaning from $s$ in other sections. Furthermore, for the true parameters, we denote $B^*=[\tvB{}^{(1)},\cdots,\tvB{}^{(K)}]$ as the matrix whose columns are $\tvB{}^\k$. Besides, we use $\tB_j$ to denote the $j$-th row of $\tB$. 

\textbf{Assumptions and definitions.} Consider the following list of assumptions and definitions:
\begin{enumerate}[leftmargin=*]
    \item There exists $\gamma\in(0,1]$ such that $\norm{A}_\infty \leq 1-\gamma $, where $A_{js}=\max_{k\in[K]}\abs{\rbr{\Sigma_{S^cS}^\k(\Sigma_{SS}^\k)^{-1}}_{js} }$ for $j\in S^c$ and $s\in S$.
    \item  There exist constants $0<\Lambda_\tmin \leq \Lambda_\tmax < \infty$ such that all eigenvalues of $\Sigma_{SS}^\k$ are in $[\Lambda_\tmin, \Lambda_\tmax]$ for all $k=1,2,\cdots,K$.
    % \item There exists a constant $D_\tmax < \infty$ such that $\max_{k\in[K]}\norm{(\Sigma_{SS}^\k)^{-1}}_\infty \leq D_\tmax$.
    \item $\rho_u: = \max_{j\in S^c, k\in[K]}\rbr{\Sigma_{S^c S^c | S}^\k}_{jj}$
    \item $\sigma_\tmax := \max_{k\in[K]} \sigma^\k$
    \item $b_\tmin := \min_{j\in S}\norm{\tB_j}_2/\sqrt{K}$.
\end{enumerate}

With the above assumptions, we are ready to present the theorem.

\begin{theorem}
\label{thm:glasso-linear-support} Assume the problem setting and assumptions in this section stated above. 
Consider the following \textbf{$l_1/l_2$-regularized Lasso problem}:
\begin{align}
    \min_{B\in\R^{p\times K}} \frac{1}{2n}\sum_{k=1}^K \|\vY^\k-\Xk\vB^\k\|_2^2 + \lambda \|B \|_{l_1/l_2},
    \label{eq:joint-estimator-linear}
\end{align}
where $B = [\vB^{(1)}, \cdots, \vB^\k]$. If the following condition holds
\begin{align*}
    n \geq \kappa_6 s \log p,\\
    K \leq c_0 \log p\\
    \sqrt{ \frac{8\sigma_\tmax^2  \log p}{\Lambda_\tmin n} } + \frac{2 }{\Lambda_\tmin}\sqrt{\frac{sp\log p}{nK}} = o\rbr{b^*_\tmin},
\end{align*}
then \eqref{eq:joint-estimator-linear} has a unique solution $\hB$, and that with probability at least
\begin{align*}
    1 - c_1 K\exp\rbr{-c_2(n-s)}- \exp\rbr{-c_3 \log p}-  s\exp\rbr{-c_4K\log p},
\end{align*}
the support union of $\hB$ is the same as $S$, and that
$
    \norm{\hB - \tB}_{l_\infty/l_2}/\sqrt{K} = o(b_\tmin)$.
    
In this statement, 
$\kappa_6$ is a constant depending on $\gamma,\Lambda_\tmin,\rho_u,\sigma_\tmax$ and $c_i$ are universal constants.

\end{theorem}

\subsection{Proof of Theorem~\ref{thm:glasso-linear-support}}

The proof is based on a constructive procedure as specified by Lemma~\ref{lm:primal-dual-pair}, which characterizes an optimal primal-dual pair for which the primal solution $\hB$ correctly recovers the support set $S$.
\begin{lemma}
\label{lm:primal-dual-pair}
Define a pair $\hB=[\hvB^{(1)},\cdots,\hvB^{(K)}]$ and $\hZ=[\hvZ^{(1)},\cdots, \hvZ^{(K)}]$ as follows.
\begin{subequations}
\begin{align}
    \hB_{S^c}&:= 0\\
    \hB_S&:= \argmin_{B_S\in\R^{s\times K}}  \frac{1}{2n}\sum_{k=1}^K \|\vY^\k-\Xk_S \vB^\k_S\|_2^2 + \lambda \|B_S \|_{l_1/l_2}
    \label{eq:socp-B}
    \\
    \hvZ_S^\k & := -\lambda^{-1}\rbr{\hSigma_{SS}^\k(\hvB_S^\k - \beta_S^{*(k)}) - \frac{1}{n}\Xk_S{}^\top \vW^\k}
    \label{eq:socp-Z}\\
    \hvZ_{S^c}^\k & := -\lambda^{-1}\rbr{\hSigma_{S^c S}^\k(\hvB_S^\k - \beta_S^{*(k)}) - \frac{1}{n}\Xk_{S^c}{}^\top \vW^\k}
    \label{eq:socp-Zc}
\end{align}
\label{eq:def-hBZ}
\end{subequations}

The following statements hold true.
\begin{enumerate}[leftmargin=*]
    \item[(a)] If the matrix $\hZ_{S^c}:=[\hvZ_{S^c}^{(1)},\cdots,\hvZ_{S^c}^{(K)}]$ defined by \eqref{eq:socp-Zc} satisfies
    \begin{align}
        \label{eq:cond-no-false}
        \|\hZ_{S^c}\|_{l_\infty/l_2} < 1,
    \end{align}
    then $(\hB, \hZ)$ is a primal-dual optimal solution to the $l_1/l_2$-regularized Lasso problem in \eqref{eq:joint-estimator-linear}. Furthermore, any optimal solution $\hB$ to \eqref{eq:joint-estimator-linear} satisfies $\hB_{S^c}=\vzero$. 
    %If $\hSigma_{SS}^\k\succ 0 $, then $\hB$ is the unique optimal primal solution.1
    \item[(b)] Define a matrix $U_S=[\vU_S^{(1)},\cdots,\vU_S^{(K)}]$ whose column vectors are 
    \begin{align*}
        \vU_S^\k:= \hvB_S^\k - \beta_S^{*\k} = (\hSigma_{SS}^\k)^{-1}\rbr{\frac{1}{n}X_S^\k{}^\top \vW^\k -\lambda \hvZ_S^\k }.
    \end{align*}
    If the conditions in (a) are satisfied, and furthermore, $U_S$ satisfies
    \begin{align}
        \label{eq:cond-no-exclude}
        \frac{\|U_S\|_{l_\infty/l_2}}{\sqrt{K}}\leq \frac{1}{2}b^*_\tmin,
    \end{align}
    then $\hB$ correctly recovers the union support $S$. That is,
    \begin{align*}
        \cbr{i\in[p]: \exists k \in [K]\,s.t.\, \hvB^\k_i\neq 0 } = S.
    \end{align*}
\end{enumerate}

\end{lemma}

\begin{remark}
Note that the matrix $\hZ_S:=[\hvZ_S^{(1)},\cdots,\hvZ_S^{(K)}]$ defined by \eqref{eq:socp-Z} is a dual solution to the restricted optimization in \eqref{eq:socp-B}, and thus satisfies
$\hZ_S\in \partial\|\hB_S\|_{l_1/l_2}$.
\end{remark}

\begin{proof}
The proof of Lemma~\ref{lm:primal-dual-pair} (a) is similar to Lemma~1 in \citep{wang2014collective} and Lemma~2 in \citep{obozinski2011support}. The proof of Lemma~\ref{lm:primal-dual-pair} (b) is straightforward from the condition in \eqref{eq:cond-no-exclude}. By definition of $b_\tmin$, \eqref{eq:cond-no-exclude} implies $\|\hvB_j^\allk\|_2 \geq \|\tvB_j{}^\allk\|_2 - \| \hvB_j^\k - \beta_j^{*\k}\|_2 \geq \frac{1}{2}\sqrt{K}b_\tmin^* >0$ for any $j\in S$.
\end{proof}

Based on Lemma~\ref{lm:primal-dual-pair}, if we can show the primal-dual pair defined in its statement can satisfy both conditions in \eqref{eq:cond-no-false} and \eqref{eq:cond-no-exclude}, then the support recovery guarantee is proved. We provide the analysis of these two conditions in Appendix~\ref{sec:no-false-recovery} and Appendix~\ref{sec:no-exclude} respectively.

Collecting the results in Appendix~\ref{sec:no-false-recovery} and Appendix~\ref{sec:no-exclude}, we conclude that, if the following conditions are satisfied:
\begin{align*}
    n \geq \kappa_6 s \log p, \quad 
    K \leq \frac{5}{64}\log p,\\
    \sqrt{ \frac{8\sigma_\tmax^2  \log p}{\Lambda_\tmin n} } + \frac{2 }{\Lambda_\tmin}\sqrt{\frac{sp\log p}{nK}} = o\rbr{b^*_\tmin},
\end{align*}
then with probability at least
\begin{align*}
    &1-3K\exp\rbr{-\frac{n}{2}\rbr{\frac{1}{4}-\sqrt{\frac{s}{n}}}_+^2} - K\exp\rbr{-(1+\delta)(n-s)\sbr{1- 2\sqrt{\frac{1}{1+\delta}}}} \\
    &- \exp\rbr{-2\frac{3}{4} \log p}-  s\exp\rbr{-2K\log p\rbr{1- 2\sqrt{\frac{1}{2\log p}}}}\\
    &\geq 1 - c_1 K\exp\rbr{-c_2(n-s)}- \exp\rbr{-c_3 \log p}-  s\exp\rbr{-c_4K\log p},
\end{align*}
conditions in \eqref{eq:cond-no-false} and \eqref{eq:cond-no-exclude} are satisfied and therefore $\hB$ correctly recovers the union support $S$.

\subsubsection{No false recovery: $\|\hZ_{S^c}\|_{l_\infty/l_2} < 1$}
\label{sec:no-false-recovery}
% \begin{lemma}[No false recovery]
% \label{lm:no-false-recovery}
% Define $(\hB,\hZ)$ by \eqref{eq:def-hBZ}. Then
% \begin{align}
%     \Pr[\|\hZ_{S^c}\|_{l_\infty/l_2} < 1] \geq 
% \end{align}
% \end{lemma}

Denote the row vectors of $\hZ_{S^c}$ as $\hZ_j := [\hvZ_j^{(1)},\cdots, \hvZ_j^{(k)}]$. Then 
\begin{align*}
    \|\hZ_{S^c}\|_{l_\infty/l_2} = \max_{j\in S^c} \|\hZ_j\|_2.
\end{align*}
Since
\begin{align*}
    \hZ_j =& \underbrace{\E[\hZ_j \mid \mX_S^\allk ]}_{T_{j1}} + \underbrace{\E[\hZ_j \mid \mX_S^\allk, \vW^\allk ]-\E[\hZ_j \mid \mX_S^\allk ]}_{T_{j2}}\\
    &+ \underbrace{\hZ_j - \E[\hZ_j \mid \mX_S^\allk, \vW^\allk ]}_{T_{j3}},
\end{align*}
to prove $\|\hZ_{S^c}\|_{l_\infty/l_2} < 1$, we resort to bound $\max_{j\in S^c} \|T_{ja}\|_2$ for $a =1,2,3$ separately. The analyses of $T_{j1}$ and $T_{j2}$ largely follow the arguments in \citep{wang2014collective} and \citep{obozinski2011support}, so details are omitted for brevity. We summarize the results of these two terms below, after which we present the detailed analysis for $T_{j3}$. 

\subsubsection{Analysis of $T_{j1}$ and $T_{j2}$}
$T_{j1}$: Following the same arguments as the derivations of Equation (28) in \citep{wang2014collective} and Equation (39) in \citep{obozinski2011support}, we have
$
   \max_{j\in S^c }\|T_{j1}\|_2 \leq 1 - \gamma$.

$T_{j2}$: Following the same arguments as the derivations of Equation (32) in \citep{wang2014collective}, we have that 
\begin{align*}
    \max_{j\in S^c}\|T_{j2}\|_2 \leq (1-\gamma ) \|\hZ_S - Z_S^* \|_{l_\infty/l_2} + (1-\gamma)\E[\|\hZ_S - Z_S^* \|_{l_\infty/l_2} \mid \mX_S^\allk],
\end{align*}
where the rows of $Z^*_S$ are defined as $Z^*_i:=\tB_i / \|\tB_i\|$ for $i\in S$. Define the matrix $\Delta\in \R^{s\times K}$ with rows $\Delta_i:=(\hB_i- \tB_i) / \|\tB_i\|_2$. By Lemma~\ref{lm:Z-diff}, if 
 $\|\Delta\|_{l_\infty/l_2} < 1/2$, then
it holds true that
\begin{align*}
    \max_{j\in S^c}\|T_{j2}\|_2 \leq 4(1-\gamma)\rbr{ \|\Delta\|_{l_\infty/l_2} + \E[\|\Delta \|_{l_\infty/l_2} \mid \mX_S^\allk] }.
\end{align*}
We will show later in the analysis pf $U_S$ that $\|\Delta \|_{l_\infty/l_2}$ is of order $o(1)$ with high probability. 
% ======
% The following conditional independencies will be useful in the analysis:
% \begin{align}
%     \vw^\k \perp \Xk_{S^c} \mid X^\allk_S,
%     \quad \hvZ^\k_S \perp   \Xk_{S^c} \mid X^\allk_S,
%     \quad \hvZ^\k_S \perp \Xk_{S^c} \mid \cbr{X^\allk_S, \vw^\allk }.
%     \label{eq:c-independence}
% \end{align}
% Besides, the following equation for $\hvZ_j^\k$ for $j\in S^c$ will also be useful:
% \begin{align*}
%     \hvZ_{j}^\k & = -\lambda^{-1}\rbr{\hSigma_{j S}^\k(\hSigma_{SS}^\k)^{-1}\rbr{\frac{1}{n}X_S^\k{}^\top \vW^\k -\lambda \hvZ_S^\k } - \frac{1}{n}\Xk_{j}{}^\top \vW^\k}\\
%     & = -\frac{1}{\lambda n}\Xk_j{}^\top(\Pi_S^\k - I_n) \vW^\k + \frac{1}{n} X_{S^c}^\k{}^\top X_S^\k (\hSigma_{SS}^\k)^{-1}\hvZ_{S}^\k,
% \end{align*}
% where $\Pi_S^\k = \frac{1}{n} \Xk_S(\hSigma_{SS}^\k)^{-1}\Xk_S{}^\top$.
% This equation is obtained by substituting $\hvB_S^\k - \beta_S^{*\k} = (\hSigma_{SS}^\k)^{-1}\rbr{\frac{1}{n}X_S^\k{}^\top \vW^\k -\lambda \hvZ_S^\k }$ into \eqref{eq:socp-Zc}.

\subsubsection{Analysis of $T_{j3}$}

Following the same arguments as the derivations of Equation (36) in \citep{wang2014collective} and Equation (42) \citep{obozinski2011support}, we have that for each $j\in S^c$,
\begin{align*}
   \text{given }\rbr{\mX_S^\allk, \vW^\allk},\\
   \hvZ_{j}^\k  - \E[\hvZ_{j}^\k\mid \mX_S^\allk, \vW^\allk] & \overset{d.}{=} \sigma_{jk} \xi_{jk},
\end{align*}
where
\begin{align*}
\begin{cases}
\xi_{jk} \sim \gN(0,1),\\ 
    \sigma_{jk}^2  := (\Sigma_{S^c S^c | S}^\k)_{jj}M_k \leq \rho_u M_k,\\
    M_k  := \frac{1}{n} \hvZ_S^\k{}^\top (\hSigma_{SS}^\k)^{-1}, \hvZ_S^\k - \frac{1}{n^2\lambda^2} \vw^\k{}^\top (\Pi_S^\k - I_n) \vw^\k.
\end{cases}
\end{align*}
Therefore,
\begin{align}
    \text{given }\rbr{\mX_S^\allk, \vW^\allk}, \nonumber \\
  \max_{j\in S^c} \|\hvZ_{j}^\k  - \E[\hvZ_{j}^\k\mid \mX_S^\allk, \vW^\allk]\|_2^2 & \overset{d.}{=}  \max_{j\in S^c} \sum_{k=1}^K \sigma_{jk}^2 \xi_{jk}^2 \nonumber \\
  & \leq \rho_u \max_{k\in[K]}|M_k| \max_{j\in S^c}\sum_{k=1}^K\xi_{jk}^2 
  \label{eq:T3-bound}
\end{align}

\textit{(1) Bounding $\max_{k\in[K]}|M_k|$.}

Bound the term $\frac{1}{n} \hvZ_S^\k{}^\top (\hSigma_{SS}^\k)^{-1}, \hvZ_S^\k$ is based on the following relations:
\begin{align*}
    \max_{k\in[K]}\|\vz_S^{*\k}\|_2 &\leq \sqrt{s} \quad \text{(by definition)},\\
    \max_{k\in[K]}\|(\hSigma_{SS}^\k)^{-1}\|_2 &\leq \frac{2}{\Lambda_\tmin}\quad \text{w.p.}\geq 1-K\exp\rbr{-\frac{n}{2}\rbr{\frac{1}{4}-\sqrt{\frac{s}{n}}}_+^2}\quad \text{(Lemma 10 in \citep{obozinski2011support})}. 
\end{align*}
Therefore, with probability $\geq 1-K\exp\rbr{-\frac{n}{2}\rbr{\frac{1}{4}-\sqrt{\frac{s}{n}}}_+^2}$,
\begin{align*}
    \frac{1}{n}\abs{ \hvZ_S^\k{}^\top (\hSigma_{SS}^\k)^{-1} \hvZ_S^\k } \leq \frac{1}{n} \|(\hSigma_{SS}^\k)^{-1}\|_2 \| \hvZ_S^\k\|_2^2 \leq \frac{1}{n} \frac{2s}{\Lambda_\tmin},
\end{align*}
For the second term in $\max_{k\in[K]}|M_k|$, note that
\begin{align*}
    \vw^\k{}^\top (I_n - \Pi_S^\k) \vw^\k & \deq \sigma^{\k 2}\sum_{j=1}^{n-s}\zeta_{jk}^2 \quad \text{with }\zeta_{jk}\sim \gN(0,1).
\end{align*}
By Lemma~\ref{lm:con-chi}, for any $\delta > 0$, 
\begin{align*}
    \Pr\sbr{\max_{k\in[K]}\sum_{j=1}^{n-s}\zeta_{jk}^2\geq 2(1+\delta)(n-s) }\leq K\exp\rbr{-(1+\delta)(n-s)\sbr{1- 2\sqrt{\frac{1}{1+\delta}}}}.
\end{align*}
To summarize, with probability at least
\begin{align*}
    1-K\exp\rbr{-\frac{n}{2}\rbr{\frac{1}{4}-\sqrt{\frac{s}{n}}}_+^2} - K\exp\rbr{-(1+\delta)(n-s)\sbr{1- 2\sqrt{\frac{1}{1+\delta}}}},
\end{align*}
it holds that 
\begin{align*}
    \max_{k\in[K]}|M_k| < \frac{1}{n}\frac{2s}{\Lambda_\tmin} + \frac{2\sigma_\tmax^2(1+\delta) (n-s) }{n^2\lambda^2}.
\end{align*}

\textit{(2) Bounding $\max_{j\in S^c}\sum_{k=1}^K\xi_{jk}^2 $}.

Combining the bound on $\max_{k\in[K]}|M_k|$ with \eqref{eq:T3-bound}, it implies
\begin{align*}
    &\max_{j\in S^c}\|T_{j3}\|_2^2 \leq \rho_u \rbr{\frac{1}{n}\frac{2s}{\Lambda_\tmin} + \frac{2\sigma_\tmax^2(1+\delta) (n-s) }{n^2\lambda^2} } \max_{j\in S^c}\sum_{k=1}^K\xi_{jk}^2\\
    &\Longrightarrow
    \cbr{\max_{j\in S^c}\|T_{j3}\|_2 < \gamma}\subseteq \cbr{ \max_{j\in S^c}\sum_{k=1}^K\xi_{jk}^2 < \frac{\gamma^2}{2\rho_u} \frac{\lambda^2n}{\lambda^2 s/\Lambda_\tmin + \sigma_\tmax^2(1+\delta)\frac{n-s}{n} }}.
\end{align*}

What's left is to bound the term $\sum_{k=1}^K\xi_{jk}^2 $. Take $t^* = \frac{\gamma^2}{4\rho_u} \frac{\lambda^2n}{\lambda^2 s/\Lambda_\tmin + \sigma_\tmax^2(1+\delta) \frac{n-s}{n}}$. If $t^*>K$, 
by Lemma~\ref{lm:con-chi} and a union bound over $j\in S^c$, we have that
\begin{align*}
    \Pr[\max_{j\in S^c}\sum_{k=1}^K\xi_{jk}^2 \geq 2t^*] \leq (p-s)\exp\rbr{-t^*\sbr{1-2\sqrt{\frac{K}{t^*}}}}.
\end{align*}

\textit{(3) Collecting all results.}

To conclude, for any $\delta>0$, if the following conditions are satisfied:
\begin{align}
    &\|\Delta\|_{l_\infty/l_2}<\frac{1}{2}, \nonumber \\
    &t^* = \frac{\gamma^2}{4\rho_u} \frac{\lambda^2n}{\lambda^2 s/\Lambda_\tmin + \sigma_\tmax^2(1+\delta)\frac{n-s}{n} } > K, \label{eq:cond-k-t}
\end{align}
then for any $\delta>0$,  with probability at least
\begin{align*}
    1&-K\exp\rbr{-\frac{n}{2}\rbr{\frac{1}{4}-\sqrt{\frac{s}{n}}}_+^2} - K\exp\rbr{-(1+\delta)(n-s)\sbr{1- 2\sqrt{\frac{1}{1+\delta}}}} \\
    &- (p-s)\exp\rbr{-t^*\sbr{1-2\sqrt{\frac{K}{t^*}}}},
\end{align*}
it holds that 
\begin{align*}
    \max_{j\in S^c }\|T_{j3}\|_2 < \gamma.
\end{align*}

\textit{(4) Condition in \eqref{eq:cond-k-t}.}

If we assume that 
\begin{align*}
    n \geq C s \log p
\end{align*}
for some constant $C$. Then
\begin{align*}
    t^* \geq \frac{\gamma^2}{4\rho_u} \frac{\lambda^2n}{\lambda^2 n/(C\Lambda_\tmin \log p) + \sigma_\tmax^2(1+\delta)\frac{n-s}{n} } = \frac{\gamma^2}{4\rho_u} \frac{1}{(C\Lambda_\tmin \log p)^{-1} + \sigma_\tmax^2(1+\delta)\frac{n-s}{\lambda^2n^2} }.
\end{align*}
With the specified choice of parameter $\lambda = \sqrt{p\log p / n}$, it implies
\begin{align*}
    t^* \geq \frac{\gamma^2}{4\rho_u} \frac{\log p}{(C\Lambda_\tmin )^{-1} + \sigma_\tmax^2(1+\delta)\frac{n-s}{np} }.
\end{align*}
Assume $C$ is chosen such that $C\geq \Lambda_\tmin^{-1}\rbr{\frac{\gamma^2}{20\rho_u} - \sigma_\tmax^2(1+\delta) \frac{n-s}{np} }{}^{-1}$ which can be easily satisfied since $\frac{n-s}{np} <1$. Then it implies $t^* \geq 5\log p$. To satisfy the condition in \eqref{eq:cond-k-t}, it is sufficient to assume $
    K \leq \frac{5}{64}\log p$,
which implies $K \leq \frac{1}{64}t^* $. Furthermore, it implies
$
    \exp\rbr{-t^*\sbr{1-2\sqrt{\frac{K}{t^*}}}} < \exp\rbr{-3\frac{3}{4}\log p}$.
    
To conclude, if $ \|\Delta\|_{l_\infty/l_2}<\frac{1}{2}$ and that
\begin{align*}
    n \geq \kappa_6 s \log p, \quad 
    K \leq \frac{5}{64}\log p,
\end{align*}
then $\max_{j\in S^c }\|T_{j3}\|_2 < \gamma$ holdes with probability at least
\begin{align*}
    1&-K\exp\rbr{-\frac{n}{2}\rbr{\frac{1}{4}-\sqrt{\frac{s}{n}}}_+^2} - K\exp\rbr{-(1+\delta)(n-s)\sbr{1- 2\sqrt{\frac{1}{1+\delta}}}} \\
    &- (p-s)\exp\rbr{-3\frac{3}{4} \log p}.
\end{align*}

\subsubsection{No exclusion: $\frac{\|U_S\|_{l_\infty/l_2}}{\sqrt{K}}\leq \frac{1}{2}b^*_\tmin$}
\label{sec:no-exclude}

\eqref{eq:socp-Z} implies that
\begin{align*}
  \hvB^\k_S - \beta^{*(k)}_S = \rbr{\hSigma^\k_{SS}}^{-1} \rbr{\frac{1}{n}\Xk_S{}^\top \vW^\k -  \lambda\hvZ^\k_S}.
\end{align*}
Define $\bw^\k := \frac{1}{\sqrt{n}} (\hSigma^\k_{SS})^{-1/2} \Xk_S{}^\top \vW^\k \deq \sigma^\k \mathbb{\xi}_k$ with $\mathbb{\xi}_k \sim \gN(0,I_p)$. Then
\begin{align*}
   \hvB^\k_S - \beta^{*(k)}_S \deq \underbrace{(\hSigma_{SS}^\k)^{-1/2}\frac{\bw^\k}{\sqrt{n}}}_{A^\k } -  \underbrace{(\hSigma_{SS}^\k)^{-1} \lambda\hvZ_{S}^\k}_{B^\k}.
\end{align*}
Denote the $i$-the entry in the vector $A^\k$ as $A^\k_i$. Then for a fixed $i \in S$, the entry $\{A^\k_i\}_{k\in[K]}$ are independent. Its easy to see that
\begin{align*}
    A^\k_i \mid X^\allk \deq \frac{\sigma^\k}{\sqrt{n}}\sqrt{\rbr{ (\hSigma^k_{SS})^{-1} }_{ii}} \xi_{ik}\quad \text{with }\xi_{ik}\sim \gN(0,1)\text{ and Cov}(\xi_{ik},\xi_{ik'})=0.
\end{align*}
Therefore,
\begin{align*}
    \max_{i\in S}\sum_{k=1}^K A^\k_i{}^2 \mid X^\allk \leq \frac{\sigma_\tmax^2}{n}\max_{k\in[K]} \norm{ (\hSigma_{SS}^\k)^{-1} }_2 \max_{i\in S}\sum_{k=1}^K \xi_{ik}^2.
\end{align*}
Since we have
\begin{align*}
    \Pr\sbr{\max_{k\in[K]} \norm{ (\hSigma_{SS}^\k)^{-1} }_2 \leq \frac{2}{\Lambda_\tmin}} \geq 1- K \exp\rbr{-\frac{n}{2}\rbr{\frac{1}{4}-\sqrt{\frac{s}{n}}}_+^2} \quad \text{by Lemma 10 in \citep{obozinski2011support}},\\
    \Pr\sbr{\max_{i\in S}\sum_{k=1}^K \xi_{ik}^2 \leq 4K\log p } \geq 1- s\exp\rbr{-2K\log p\rbr{1- 2\sqrt{\frac{1}{2\log p}}}}\,\, \text{by Lemma~\ref{lm:con-chi}},
\end{align*}
then with probability at least
$
    1- K \exp\rbr{-\frac{n}{2}\rbr{\frac{1}{4}-\sqrt{\frac{s}{n}}}_+^2} -  s\exp\rbr{-2K\log p\rbr{1- 2\sqrt{\frac{1}{2\log p}}}}$,
it holds that
\begin{align*}
    \max_{i\in S}\sqrt{\sum_{k=1}^K A^\k_i{}^2} \leq \sqrt{ \frac{8\sigma_\tmax^2 K \log p}{\Lambda_\tmin n} }.
\end{align*}
Tuning now to the term $B^\k$: 
\begin{align*}
   &  \max_{i\in S}\sqrt{\sum_{k=1}^K B^\k_i{}^2} = \lambda \max_{i\in S}\sqrt{ \sum_{k=1}^K \rbr{\ve_i^\top (\hSigma_{SS}^\k)^{-1}\hvZ_{S}^\k}^2}\\
   &\leq  \lambda\max_{i\in S}\sqrt{\sum_{k=1}^K \|(\hSigma_{SS}^\k)^{-T}\ve_i\|_2^2\|\hvZ_{S}^\k\|_2^2} \quad \text{by Cauchy-Schwarz inequality} \\
   & \leq \lambda \max_{i\in S}\max_{k\in[K]} \|(\hSigma_{SS}^\k)^{-T}\ve_i\|_2\sqrt{\sum_{k=1}^K \|\hvZ_{S}^\k\|_2^2} \leq \lambda \max_{k\in[K]}\|(\hSigma_{SS}^\k)^{-1}\|_2 \sqrt{s}.
\end{align*}
The last inequality holds because $\|\hZ_S\|_{l_\infty/l_2}\leq 1$.
Applying Lemma 10 in \citep{obozinski2011support} to $\max_{k\in[K]}\|(\hSigma_{SS}^\k)^{-1}\|_2$ again, with probability at least $1- K \exp\rbr{-\frac{n}{2}\rbr{\frac{1}{4}-\sqrt{\frac{s}{n}}}_+^2}$, it holds that 
\begin{align*}
    \textstyle &\max_{i\in S}\sqrt{\sum_{k=1}^K B^\k_i{}^2} \leq \frac{2\lambda \sqrt{s}}{\Lambda_\tmin}.
\end{align*}

To conclude, with probability at least
\begin{align}
     1- 2K \exp\rbr{-\frac{n}{2}\rbr{\frac{1}{4}-\sqrt{\frac{s}{n}}}_+^2} -  s\exp\rbr{-2K\log p\rbr{1- 2\sqrt{\frac{1}{2\log p}}}},
     \label{eq:prob-bb}
\end{align}
it holds that (with specified $\lambda = \sqrt{\frac{p\log p}{n}}$)
\begin{align*}
    &\|\hB_S - \tB_S\|_{l_\infty/l_2} \leq  \sqrt{ \frac{8\sigma_\tmax^2 K \log p}{\Lambda_\tmin n} } + \frac{2 }{\Lambda_\tmin}\sqrt{\frac{sp\log p}{n}}\\
    &\Longrightarrow \frac{\|U_S\|_{l_\infty/l_2}}{\sqrt{K}} \leq  \sqrt{ \frac{8\sigma_\tmax^2  \log p}{\Lambda_\tmin n} } + \frac{2 }{\Lambda_\tmin}\sqrt{\frac{sp\log p}{nK}}.
\end{align*}
Therefore, if 
\begin{align*}
    \sqrt{ \frac{8\sigma_\tmax^2  \log p}{\Lambda_\tmin n} } + \frac{2 }{\Lambda_\tmin}\sqrt{\frac{sp\log p}{nK}} \leq \frac{1}{2} b^*_\tmin,
\end{align*}
then \eqref{eq:cond-no-exclude} is satisfied with probability specified in \eqref{eq:prob-bb}.

\section{Invariant Sets}
\label{app:invariant}

We often need to take a union control over all permutations $\pi\in\sS_p$ in the proofs. However, in these steps, we often care about the connection matrices $\tG^\allk(\pi)$ instead of the permutation $\pi$ itself. Therefore, we want to see whether we can control a fewer number of events instead of enumerating over all $p!$ many permutations. Alternatively, we want to should that, given $\Sigma^\allk$, the number of elements in the set $\{\tG^\allk(\pi): \pi\in\sS_p\}$ can be fewer than $p!$. %Note that \citep{aragam2017learning} has shown for the special case when $K=1$ that $\{\tG^\allk(\pi): \pi\in\sS_p\}$ has at most $p{p\choose d}$ many elements. Therefore, the arguments in this section are inspired by \citep{aragam2017learning}.

We start with the following definition which specifies the population-level quantity that we are interested in.
\begin{definition}[Population SEM] For any $S\subseteq[p]\setminus\{j\}$, let 
    \begin{align}
        g_j^\k(S):= \argmin_{g\in\R^p,\text{supp}(g)\subseteq S}\E[\rvX_j^\k - g^\top \rvX^\k]^2,
    \end{align}
    where $\rvX^\k$ is the random variable that follows $\gN(0,\Sigma^\k)$.
\end{definition}

$ g_j^\k(S)$ is called the SEM coefficients for variable $\rvX_j$ regressed on the nodes in $S$~\citep{aragam2017learning,aragam2019globally}. It is a population-level quantity that depends on $\Sigma^\k$, but not on the sample $\mX^\k$. In \citep{aragam2017learning,aragam2019globally}, this quantity is used for a similar purpose on the single DAG estimation task. It is easy to verify that $g_j^\k(S_j(\pi)) = \tvG_j^\k(\pi)$. Lemma~\ref{lm:invariant} summarizes the key observations.

\begin{lemma} \label{lm:invariant}
Let $S_j(\pi)=\{i: \pi(i)<\pi(j)\}$ and $\tvG_j^\k(\pi)$ is the $j$-th column of $\tG^\k(\pi)$. Then
\begin{align*}
    g_j^\k(S)=\tvG_j^\k(\pi)
\end{align*}
for any set $S$ such that
\begin{align}
    \text{supp}\big(\tvG_j^\k(\pi)\big)\subseteq S\subseteq S_j(\pi) \label{eq:invariant}.
\end{align}
Since the set of union parents $U_j(\pi):=\cup_{k\in[K]}\text{supp}\big(\tvG_j^\k(\pi)\big)$ satisfies \eqref{eq:invariant}, it implies that
\begin{align}
    g_j^\k(U_j(\pi))= \tvG_j^\k(\pi),\quad \forall k\in[K].
\end{align}
\end{lemma}

A direct consequence of this Lemma~\ref{lm:invariant} is that:
\begin{align*}
    \Big|\{\tvG_j^\allk(\pi):\pi\in\sS_p\}\Big| = \Big|\{\tvG_j^\allk(U_j(\pi)):\pi\in\sS_p\}\Big| \leq \Big|\{U_j(\pi):\pi\in\sS_p\}\Big|.
\end{align*}
Recall that $\uniond_j:=\max_{\pi\in\sS_p}|U_j(\pi)|$. Then  there are at most $\sum_{ 0\leq m \leq \uniond_j } {p\choose m}$ many elements in this set. Note that 
\begin{align*}
    \sum_{0\leq m \leq \uniond_j}{p\choose m} \leq \sum_{0\leq m \leq \uniond}{p\choose m} \leq p^d.
\end{align*}
The last inequality holds for all $p\geq d\geq 2$. Therefore, 
\begin{align}
    \Big|\{\tvG_j^\allk(\pi):\pi\in\sS_p\}\Big| \leq \Big|\{U_j(\pi):\pi\in\sS_p\}\Big| \leq \sum_{0\leq m \leq \uniond_j}{p\choose m}\leq p^\uniond. \label{eq:invariant-size}
\end{align}

\section{Details of Continuous Formulation}
\label{app:continuous}

We first prove that $\hT\in\gT_p$:
\begin{lemma}
If $(\hG^\allk,\hT)$ is a pair of optimal solution to \eqref{eq:cont-opt}, then $\hT$  is in the discrete space $\gT_p$. Equivalently, if $\hT$ is an optimal solution, then there  exists a permutation $\hpi\in\sS_p$ such that 
    \begin{align}
        \hT_{ij}=\begin{cases}
        1 & \text{if }\hpi(i) < \hpi(j), \\
        0 & \text{otherwise}.
        \end{cases} \label{eq:def-T}
    \end{align}
\end{lemma}

\begin{proof}
    By the constraint $h(T)=0$ in \eqref{eq:dag-const}, the graph structure induced by the matrix $\hT$ must be acyclic. Therefore, $\hT$ represents a DAG and has an associated topological order (causal order). Denote this order by $\hpi$. What remains is to show \eqref{eq:def-T} holds true.
    
    Assume there exists an entry $(i',j')$ such that $\hpi(i')<\hpi(j')$ and $\hT_{i'j'}\neq 1$. We construct the following pair of $(T,G^\allk)$:
    \begin{align*}
    T:=
    \begin{cases}
        T_{i'j'} = 1\\
        T_{ij} = \hT_{ij} \text{ for } (i,j)\neq(i',j')
    \end{cases}
    \quad 
    G^\k:=
    \begin{cases}
        G_{i'j'}^\k = \hT_{i'j'}\cdot \hG_{i'j'}^\allk \\
        G_{ij}^\k = \hG_{ij}^\k \text{ for } (i,j)\neq(i',j')
    \end{cases}\forall k\in[K].
    \end{align*}
    It is constructed by modifying the $(i',j')$-th entries in the solution pair $(\hT,\hG^\allk)$. It is easy to see that: (i) this constructed pair $(T,G^\allk)$ is a feasible solution to \eqref{eq:cont-opt}; and (ii) $(T,G^\allk)$ achieves a smaller objective value than $(\hT,\hG^\allk)$.
    
    The reason for (ii) is that
    after the modification, the matrix $\oG^\allk$ remains unchanged. That is, $\hT\circ \hG^\k = T \circ G^\k$. Therefore, the squared loss and the group-norm in the objective remain unchanged. However, the term $\|\vone_{p\times p} - T\|_F^2$ has been reduced by setting $T_{i'j'}=1$. 
    
    This makes a contradiction to the optimality of $(\hT,\hG^\allk)$. Therefore, the assumption is not true and we conclude that:
    \begin{align*}
        \hpi(i) < \hpi(j) \Longrightarrow \hT_{ij}=1.
    \end{align*}
    Finally, since $\hT$ is consistent with $\hpi$, by definition, $\hT_{ij}=0$ if $\hpi(i)\geq \hpi(j)$. 
\end{proof}

We now start to show the equivalence between the optimization in \eqref{eq:joint-estimator} and in \eqref{eq:cont-opt}.

Firstly, the solution search spaces are the same. We have shown that $\hT\in \gT_p$. For each element in $\gT_p$, we denote it by $\hT(\pi)$ based on its associated order $\pi$. Since $\hT(\pi)$ is a \textit{dense} DAG with topological order $\pi$, it is easy to see the space $\{G\circ \hT(\pi) : G \in \R^{p\times p}\}$ includes all DAGs that are consistent with $\hpi$ and excludes any DAGs that are not. In other words, $\{G\circ \hT(\pi) : G \in \R^{p\times p}\} = \sD(\pi)$. Therefore, the solution search spaces of these two optimization problems are equivalent.

Secondly, the optimization objectives are the same. Again, since $\hT\in \gT_p$, the term $\rho\|\vone_{p\times p} - \hT\|_F^2$ is a constant with a fixed value $\frac{\rho(p-1)p}{2}$. The remaining two terms in the objective are the same as the objective in \eqref{eq:joint-estimator}.

Since both the solution search space and the optimization objectives are equivalent, these two optimizations are equivalent.

\section{Details of Synthetic Experiments in Sec~\ref{sec:syn-experiments}}
\label{app:experiments}

\subsection{Evaluation of structure prediction}
We classify the positive predictions in three types:
\begin{itemize}
  \item True Positive: predicted association exists in correct direction.
  \item Reverse: predicted association exists in opposite direction.
  \item False Positive: predicted association does not exist
\end{itemize}
Based on them, we use five metrics:
\begin{itemize}
    \item False Discovery Rate (FDR):  (reverse + false positive) / (true positive + reverse + false positive)
    \item True Positive Rate (TPR): (true positive) / (ground truth positive)
    \item False Positive Rate (FPR): (false positive) / (ground truth positive)
    \item Structure Hamming Distance (SHD): (false negative + reverse + false positive) 
    \item Number of Non-Zero (NNZ): (true positive + reverse + false positive)
\end{itemize}
\subsection{A more complete result for Fig~\ref{fig:k-acc}}
\begin{figure}[h]
    \centering
    \includegraphics[width=\textwidth]{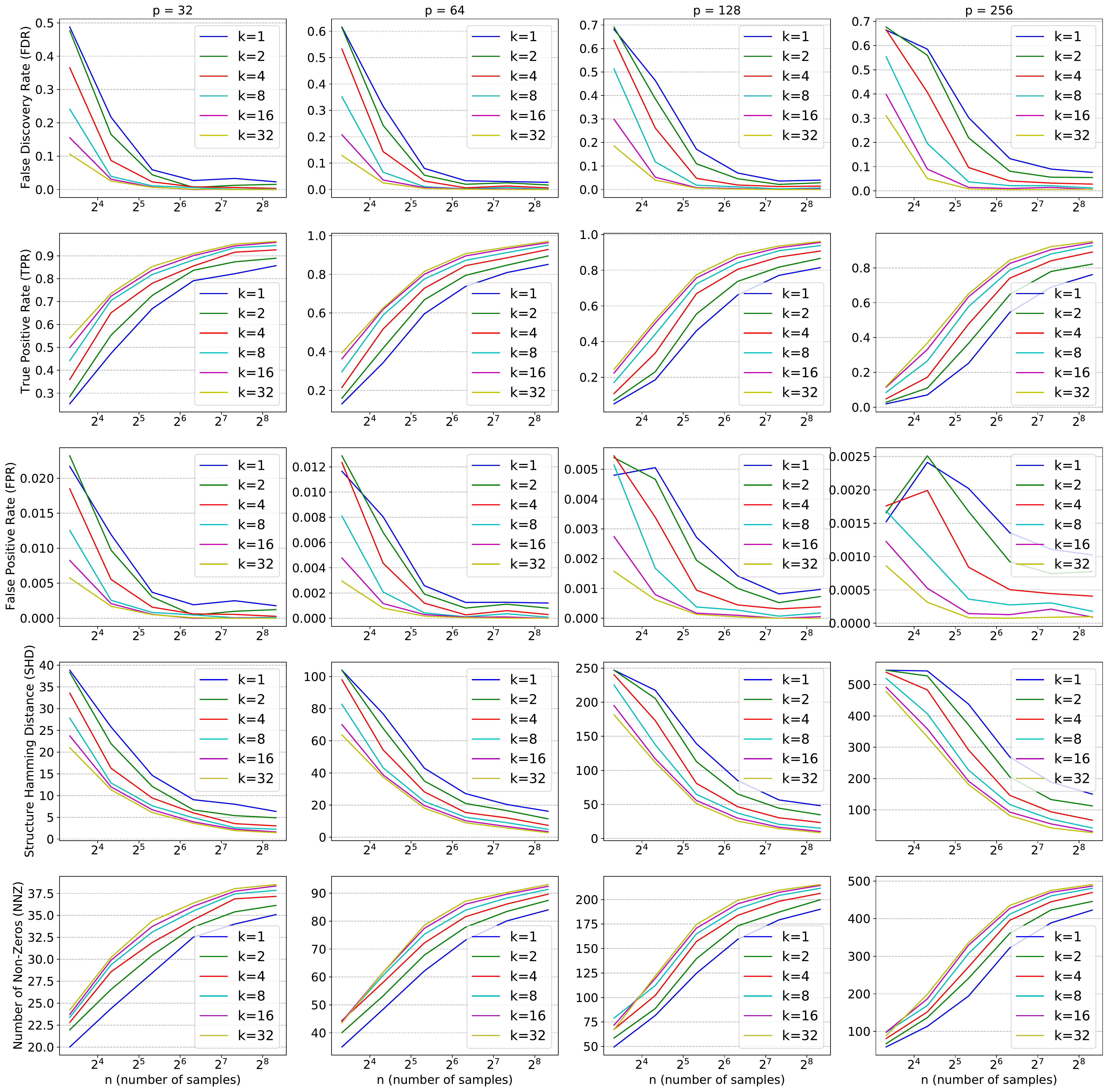}
    \caption{Full results of FDR, TPR, FPR, SHD, and NNZ.}
    \label{fig:full_acc}
\end{figure}

We demonstrate our methods on synthetic data with $(p, s) \in$ \{(32, 40), (64, 96), (128, 224), (256, 512)\}, $K \in $\{1, 2, 4, 8, 16, 32\}, $n \in $ \{10, 20, 40, 80, 160, 320\}. For each $\{p, s, K, n\}$, we run experiments on 64 graphs. We report the full results in Fig.\ref{fig:full_acc}.

\subsection{Computing resources}
Since we need to run a large set of experiments spanning different values of $\{p,s,K,n\}$, the  synthetic experiments are run on a CPU cluster containing 416 nodes. On each node, there are 24 CPUs (Xeon 6226 CPU @ 2.70GHz) with 192 GB memory. Each individual experiment is run on 4 CPUs. It takes about 10 hours to finish a complete set of experiments on about 400 CPUs.

\newpage
\section{Useful Results In Existing Works}
\begin{lemma}[Laurent-Massart]
\label{lm:chi-square}
Let $a_1,\cdots, a_m$ be nonnegative, and set 
\begin{align*}
    \|a\|_\infty = \sup_{i\in[m]} |a_i|,\quad \|a\|_2^2 =  \sum_{i=1}^m a_i^2.
\end{align*}
For i.i.d $Z_i \sim \gN(0,1)$, the following inequalities hold for any positive $t$:
\begin{align*}
    \Pr\sbr{\sum_{i=1}^m a_i(Z_i^2-1) \geq 2\|a\|_2 \sqrt{t} + 2\|a\|_\infty t }\leq e^{-t},\\
    \Pr\sbr{\sum_{i=1}^m a_i(Z_i^2-1) \leq - 2\|a\|_2 t } \leq e^{-t}.
\end{align*}
\end{lemma}

\begin{lemma}\citep{obozinski2011support}
\label{lm:con-chi}
Let $Z$ be a central Chi-squared distributed random variable with the degree $m$. Then for all $t > m$, we have
\begin{align*}
    \Pr[Z \geq 2t ] \leq \exp\rbr{-t\sbr{1-2\sqrt{\frac{m}{t}}}}.
\end{align*}

\end{lemma}

\begin{lemma}\citep{obozinski2011support}
\label{lm:Z-diff}
Consider the matrix $\Delta\in \R^{s\times K}$ with rows $\Delta_i:=(\hB_i- \tB_i) / \|\tB_i\|_2$. If $\|\Delta\|_{l_\infty/l_2} < \frac{1}{2}$, then $\|\hZ_S - Z_S^*\|_{l_\infty/l_2} \leq 4 \|\Delta\|_{l_\infty/l_2}$.
\end{lemma}

\end{document}

%% file: math_commands.tex
\usepackage{amsmath,amsfonts,bm}
\usepackage{dsfont}

\def\eqref#1{equation~\ref{#1}}
\def\1{\bm{1}}

\def\tr{{\normalfont\text{tr}}}

\def\vzero{{\bm{0}}}
\def\vone{{\bm{1}}}

\def\vve{{\bm{\varepsilon}}}

\def\ve{{\bm{e}}}

\def\vu{{\bm{u}}}

\def\vw{{\bm{w}}}

\def\vy{{\bm{y}}}
\def\vz{{\bm{z}}}

\def\mX{{\bm{X}}}

\DeclareMathAlphabet{\mathsfit}{\encodingdefault}{\sfdefault}{m}{sl}
\SetMathAlphabet{\mathsfit}{bold}{\encodingdefault}{\sfdefault}{bx}{n}

\def\tB{{\tens{B}}}

\def\tG{{\tens{G}}}

\def\gE{{\mathcal{E}}}

\def\gL{{\mathcal{L}}}

\def\gN{{\mathcal{N}}}
\def\gO{{\mathcal{O}}}

\def\gT{{\mathcal{T}}}

\def\sD{{\mathbb{D}}}
\def\sS{{\mathbb{S}}}

\newcommand{\E}{\mathbb{E}}

\newcommand{\R}{\mathbb{R}}

\DeclareMathOperator*{\argmin}{arg\,min}

\DeclareMathOperator{\sign}{sign}

\newcommand{\rbr}[1]{\left(#1\right)}
\newcommand{\sbr}[1]{\left[#1\right]}
\newcommand{\cbr}[1]{\left\{#1\right\}}

\def\bb{\begin{equation}} \def\ee{\end{equation}}

\newcommand{\abs}[1]{\left|{#1}\right|}